\documentclass[runningheads]{llncs}

\usepackage{amsmath,amssymb,graphicx,wrapfig,paralist}
\usepackage[usenames,dvipsnames]{xcolor}
\usepackage{bm}
\usepackage{algorithmicx,algorithm}
\usepackage[noend]{algpseudocode}
\usepackage{algcompatible}
\usepackage{booktabs}
\usepackage{proof}
\usepackage{tikz}
\usetikzlibrary{arrows}
\usetikzlibrary{arrows.meta}
\usepackage{tikz-uml}
\usepackage{csquotes}
\usepackage{graphics}
\usepackage{stmaryrd}
\usepackage{soul}
\usepackage{array}
\usepackage{sectsty}
\usepackage{enumitem}

\usepackage{ellipsis, mparhack, ragged2e} 
\usepackage[l2tabu, orthodox]{nag} 

\usepackage{tabu,multirow}
\usepackage{xparse}
\usepackage{mathtools}
\usepackage{environ}
\usepackage{lscape}

\usepackage{url}
\usepackage{lipsum}

\usepackage[english]{babel}
\usepackage{subcaption}
\usepackage{capt-of}

\usepackage[acronym]{glossaries}
\usepackage{algorithm}
\usepackage{algpseudocode}

\usepackage{marginfix}
\usepackage{xifthen}

\usepackage{longtable} 

\usepackage{listings}
\lstdefinestyle{customJ}{
  belowcaptionskip=1\baselineskip,
  breaklines=true,
  frame=L,
  xleftmargin=\parindent,
  language=Java,
  showstringspaces=false,
  basicstyle=\footnotesize\ttfamily,
  keywordstyle=\bfseries\color{green!40!black},
  commentstyle=\itshape\color{purple!40!black},
  identifierstyle=\color{blue},
  stringstyle=\color{orange},
}

\usepackage{afterpage}

\renewcommand{\arraystretch}{1.2}

\usepackage{xfrac}

\usepackage{hyperref}
\usepackage[capitalise]{cleveref}

\usepackage{placeins} 

\usepackage{pgfplots}
\usepgfplotslibrary{fillbetween}
\pgfplotsset{compat=newest}

\usepackage{minibox}
\usetikzlibrary{%
arrows,%
calc,%
shapes.misc,%
shapes.arrows,%
chains,%
matrix,%
positioning,%
scopes,%
decorations.pathmorphing,%
decorations.pathreplacing,%
decorations.markings,%
shadows,%
er,%
trees,%
shapes,%
automata,%
petri,%
patterns,%
shapes,%
fit,
backgrounds,
plotmarks
}
\tikzstyle{state}+=[minimum size = 6mm, inner sep=0,outer sep=1]
\tikzset{->,
	>=stealth',
	node distance=3cm, 
}
\tikzset{initial distance=2em,
every initial by arrow/.style={->}, initial
text={},accepting/.style={double distance=1.5pt,outer sep=2pt}}

\tikzstyle{none}=[]
\tikzstyle{state-split}=[ellipse, draw=black, rectangle split,
rectangle split parts=2, rectangle split draw splits=false, inner
sep=1pt, outer sep=2pt]
\tikzstyle{state}=[circle, draw=black, inner
sep=2pt, outer sep=2pt, minimum size=0.5cm]

\tikzstyle{arrow}=[-latex,draw=black] 
\tikzstyle{arrow-dashed}=[-latex,draw=black,dashed]
\tikzstyle{simple}=[-,draw=black]
\tikzstyle{simple-dashed}=[-,draw=black,dashed]

\pgfdeclarelayer{nodelayer}
\pgfdeclarelayer{edgelayer}
\pgfsetlayers{edgelayer,nodelayer,main,connections,background}

\newcommand{\distribution}{\mathsf{Dist}}

\newcommand{\mdp}{\mathcal{M}}
\newcommand{\mdpstates}{S}
\newcommand{\mdpinitstates}{s_0}
\newcommand{\mdpactions}{A}
\newcommand{\mdpavailactions}{A}
\newcommand{\mdptransitions}{P}
\newcommand{\mdppath}{\rho}
\newcommand{\mdpstrategy}{\pi}
\newcommand{\mdpfinitepaths}{\mathsf{FPaths}_\mdp}
\newcommand{\mdpinfinitepaths}{\mathsf{IPaths}_\mdp}
\newcommand{\last}{\mathsf{last}}

\newcommand{\pomdp}{\mathcal{P}}
\newcommand{\observationmap}{\mathcal{O}}
\newcommand{\observations}{Z}

\newcommand{\observationsequence}{\bar{o}}

\newcommand{\fsc}{\mathcal{F}}
\newcommand{\fscstates}{N}
\newcommand{\fscactionmap}{\gamma}
\newcommand{\fscinitstate}{n_0}
\newcommand{\fsctransitionmap}{\delta}

\newcommand{\learntable}{\mathcal{E}}
\newcommand{\upperrows}{R}
\newcommand{\columns}{C}

\DeclareGraphicsExtensions{.pdf, .png, .jpg}

\RequirePackage{marvosym}

\DeclareDocumentCommand{\post}{D<>{} O{} D(){}}{\mathsf{Post}_{#1}^{#2}\ifthenelse{\isempty{#3}}{}{(#3)}}

\newcommand{\supp}{\mathsf{supp}}

\newcommand{\N}{\mathbb{N}}

\newcommand{\tool}[1]{\textsc{#1}}

\newcommand{\learningtable}{\ensuremath{\mathcal{T}}}
\newcommand{\strategytable}{\ensuremath{\mathcal{S}}}
\newcommand{\dontcare}{\ensuremath{\dag}}
\newcommand{\dontknow}{\ensuremath{\chi}}
\newcommand{\fscset}{\ensuremath{\mathfrak{F}}}
\newcommand{\fsclearningtableinitstate}{n_{0,\learningtable}}
\newcommand{\fschypothesis}{\fsc_{hyp}}
\newcommand{\observationsequenceof}{\overline{\observationmap}}

\makeatletter
\newcommand{\algcolor}[2]{%
	\hskip-\ALG@thistlm\colorbox{#1}{\parbox{\dimexpr\linewidth-2\fboxsep}{\hskip\ALG@thistlm\relax #2}}%
}
\newcommand{\algemph}[2]{\algcolor{#1}{#2}}
\makeatother

\renewcommand{\Call}[2]{\textsc{#1}(#2)}
\renewcommand{\algorithmiccomment}[1]{\bgroup\hfill//~#1\egroup}

\definecolor{algo-teal}{rgb}{0.84, 0.85, 0.97}
\definecolor{algo-lime}{rgb}{0.898, 1, 0.53}
\definecolor{algo-pink}{rgb}{1, 0.827, 0.878}
\definecolor{algo-yellow}{rgb}{1, 0.98, 0.74}
\definecolor{persianindigo}{rgb}{0.2, 0.07, 0.48}
\definecolor{lincolngreen}{rgb}{0.11, 0.35, 0.02}
\definecolor{mulberry}{rgb}{0.77, 0.29, 0.55}
\definecolor{goldenpoppy}{rgb}{0.99, 0.76, 0.0}

\newtoggle{showplots}
\toggletrue{showplots}

\newcommand{\inlineheadingbf}[1]{\smallskip\noindent{\bfseries #1.}}

\definecolor{plotred}{RGB}{255,0,0}
\definecolor{plotgreen}{RGB}{0,255,0}
\definecolor{plotblue}{RGB}{0,0,255}
\definecolor{plotyellow}{RGB}{230,230,0}
\definecolor{plotcyan}{RGB}{0,255,255}
\definecolor{plotorange}{RGB}{255,127,0}
\definecolor{plotpink}{RGB}{255,0,255}
\definecolor{plotlightgray}{RGB}{192,192,192}
\definecolor{plotdarkgray}{RGB}{128,128,128}
\definecolor{plotdarkred}{RGB}{128,0,0}
\definecolor{plotgreenyellow}{RGB}{128,128,0}
\definecolor{plotdarkgreen}{RGB}{0,128,0}
\definecolor{plotpurple}{RGB}{128,0,128}
\definecolor{plotteal}{RGB}{0,128,128}
\definecolor{plotdarkblue}{RGB}{0,0,128}
\definecolor{plotlightred}{RGB}{205,92,92}
\definecolor{plotlightblue}{RGB}{176,196,222}
\colorlet{color1}{plotred}
\colorlet{color7}{plotgreen}
\colorlet{color3}{plotblue}
\colorlet{color2}{plotyellow}
\colorlet{color5}{plotcyan}
\colorlet{color2}{plotorange}
\colorlet{color7}{plotpink}
\colorlet{color8}{plotlightgray}
\colorlet{color9}{plotdarkgray}
\colorlet{color10}{plotdarkred}
\colorlet{color11}{plotgreenyellow}
\colorlet{color12}{plotdarkgreen}
\colorlet{color13}{plotpurple}
\colorlet{color14}{plotteal}
\colorlet{color15}{plotdarkblue}
\colorlet{color16}{plotlightred}
\colorlet{color17}{plotlightblue}

\newlength{\quantileplotwidth}
\newlength{\quantileplotheight}
\newlength\scatterplotsize
\newcommand{\scatterplotSize}[6]{%
\begin{tikzpicture}
	\begin{axis}[
		width=\scatterplotsize,
		height=\scatterplotsize,
		axis equal image,
		xmin=1,
		ymin=1,
		ymax=128,
		xmax=128,
		xmode=log,
		ymode=log,
		axis x line=bottom,
		axis y line=left,
		xtick={2,4,8,16,32,64},
		xticklabels={2,4,8,16,32,64},
		extra x ticks = {128},
		extra x tick labels = {TO},
		extra x tick style = {grid = major},
		ytick={2,4,8,16,32,64},
		yticklabels={2,4,8,16,32,64},
		extra y ticks = {128},
		extra y tick labels = {TO},
		extra y tick style = {grid = major},
		xlabel={#3},
		xlabel style={font=\scriptsize,yshift=5pt},
		ylabel={#5},
		ylabel style={font=\scriptsize,yshift=-9pt},
		yticklabel style={font=\scriptsize},
		xticklabel style={font=\scriptsize},
		legend pos=north east,
		legend columns=5,
		legend style={nodes={scale=0.75, transform shape},inner sep=1.5pt, xshift=1mm, yshift=7mm},
		set layers,
		mark layer=axis background
		]
		
		\iftoggle{showplots}{\addplot[
			scatter,
			only marks,
			scatter/classes={
				cutoffstrategy={mark=square*,color1,mark size=1.25},
				distribution={mark=triangle*,color2,mark size=1.25},
				minimizeautomaton={mark=o,color3,mark size=1.75}
			},
			scatter src=explicit symbolic
			]%
			table [col sep=tab,x=#2,y=#4,meta=heuristic
			] {#1};
		}{\node[anchor=south west, align=center, red] {\huge NOT\\ COMPILED};}
		\ifthenelse{\NOT\equal{#6}{false}}{\legend{base,H1,H2}}{}
		\addplot[no marks, name path=f] coordinates {(0.01,0.01) (128,128) };
		\addplot[name path=g] coordinates {(0.01,0.01) (128,0.01)};
		\addplot[no marks, densely dotted] coordinates {(0.01,0.1) (12.8,128)};
		\addplot[no marks, densely dotted] coordinates {(0.1,0.01) (128,12.8)};
	\end{axis}
\end{tikzpicture}
}

\newcommand{\scatterplotSizeStorm}[6]{%
	\begin{tikzpicture}
		\begin{axis}[
			width=\scatterplotsize,
			height=\scatterplotsize,
			axis equal image,
			xmin=1,
			ymin=1,
			ymax=8192,
			xmax=8192,
			xmode=log,
			ymode=log,
			axis x line=bottom,
			axis y line=left,
			extra x ticks = {8192},
			extra x tick labels = {TO},
			extra x tick style = {grid = major},
			extra y ticks = {8192},
			extra y tick labels = {TO},
			extra y tick style = {grid = major},
			xlabel={#3},
			xlabel style={font=\scriptsize,yshift=5pt},
			ylabel={#5},
			ylabel style={font=\scriptsize,yshift=-9pt},
			yticklabel style={font=\scriptsize},
			xticklabel style={font=\scriptsize},
			legend pos=north east,
			legend columns=5,
			legend style={nodes={scale=0.75, transform shape},inner sep=1.5pt, xshift=1mm, yshift=7mm},
			set layers,
			mark layer=axis background
			]
			
			\iftoggle{showplots}{\addplot[
				scatter,
				only marks,
				scatter/classes={
					cutoffstrategy={mark=square*,color1,mark size=1.25},
					distribution={mark=triangle*,color2,mark size=1.25},
					minimizeautomaton={mark=o,color3,mark size=1.75}
				},
				scatter src=explicit symbolic
				]%
				table [col sep=tab,x=#2,y=#4,meta=heuristic
				] {#1};
			}{\node[anchor=south west, align=center, red] {\huge NOT\\ COMPILED};}
			\ifthenelse{\NOT\equal{#6}{false}}{\legend{base,H1,H2}}{}
			\addplot[no marks, name path=f] coordinates {(0.01,0.01) (8192,8192) };
			\addplot[name path=g] coordinates {(0.01,0.01) (8192,0.01)};
			\addplot[no marks, densely dotted] coordinates {(0.01,0.1) (819,8192)};
			\addplot[no marks, densely dotted] coordinates {(0.1,0.01) (8192,819)};
		\end{axis}
	\end{tikzpicture}
}

\newcommand{\scatterplotTime}[6]{%
	\begin{tikzpicture}
		\begin{axis}[
			width=\scatterplotsize,
			height=\scatterplotsize,
			axis equal image,
			xmin=0.001,
			ymin=0.001,
			ymax=3600,
			xmax=3600,
			xmode=log,
			ymode=log,
			axis x line=bottom,
			axis y line=left,
			extra x ticks = {3600},
			extra x tick labels = {TO},
			extra x tick style = {grid = major},
			extra y ticks = {3600},
			extra y tick labels = {TO},
			extra y tick style = {grid = major},
			xlabel={#3},
			xlabel style={font=\scriptsize,yshift=5pt},
			ylabel={#5},
			ylabel style={font=\scriptsize,yshift=-9pt},
			yticklabel style={font=\scriptsize},
			xticklabel style={font=\scriptsize},
			legend pos=north east,
			legend columns=5,
			legend style={nodes={scale=0.75, transform shape},inner sep=1.5pt, xshift=1mm, yshift=7mm},
			set layers,
			mark layer=axis background
			]
			
			\iftoggle{showplots}{\addplot[
				scatter,
				only marks,
				scatter/classes={
					cutoffstrategy={mark=square*,color1,mark size=1.25},
					distribution={mark=triangle*,color2,mark size=1.25},
					minimizeautomaton={mark=o,color3,mark size=1.75}
				},
				scatter src=explicit symbolic
				]%
				table [col sep=tab,x=#2,y=#4,meta=heuristic
				] {#1};
			}{\node[anchor=south west, align=center, red] {\huge NOT\\ COMPILED};}
			\ifthenelse{\NOT\equal{#6}{false}}{\legend{base,H1,H2}}{}
			\addplot[no marks, name path=f] coordinates {(0.001,0.001) (3600,3600) };
			\addplot[name path=g] coordinates {(0.001,0.001) (3600,0.001)};
			\addplot[no marks, densely dotted] coordinates {(0.001,0.01) (360,3600)};
			\addplot[no marks, densely dotted] coordinates {(0.01,0.001) (3600,360)};
		\end{axis}
	\end{tikzpicture}
}

\newcommand{\scatterplotProb}[6]{%
	\begin{tikzpicture}
		\begin{axis}[
			width=\scatterplotsize,
			height=\scatterplotsize,
			axis equal image,
			xmin=0,
			ymin=0,
			ymax=1.1,
			xmax=1.1,
			xmode=linear,
			ymode=linear,
			axis x line=bottom,
			axis y line=left,
			extra x ticks = {1.1},
			extra x tick labels = {TO},
			extra x tick style = {grid = major},
			extra y ticks = {1.1},
			extra y tick labels = {TO},
			extra y tick style = {grid = major},
			xlabel={#3},
			xlabel style={font=\scriptsize,yshift=5pt},
			ylabel={#5},
			ylabel style={font=\scriptsize,yshift=-9pt},
			yticklabel style={font=\scriptsize},
			xticklabel style={font=\scriptsize},
			legend pos=north east,
			legend columns=5,
			legend style={nodes={scale=0.75, transform shape},inner sep=1.5pt, xshift=1mm, yshift=7mm},
			set layers,
			mark layer=axis background
			]
			
			\iftoggle{showplots}{\addplot[
				scatter,
				only marks,
				scatter/classes={
					cutoffstrategy={mark=square*,color1,mark size=1.25},
					distribution={mark=triangle*,color2,mark size=1.25},
					minimizeautomaton={mark=o,color3,mark size=1.75}
				},
				scatter src=explicit symbolic
				]%
				table [col sep=tab,x=#2,y=#4,meta=heuristic
				] {#1};
			}{\node[anchor=south west, align=center, red] {\huge NOT\\ COMPILED};}
			\ifthenelse{\NOT\equal{#6}{false}}{\legend{base,H1,H2}}{}
			\addplot[no marks] coordinates {(0,0) (1.1,1.1) };
			\addplot[name path=g] coordinates {(0,1.1) (1.1,1.1)};
		\end{axis}
	\end{tikzpicture}
}

\newcommand{\scatterplotRmax}[6]{%
	\begin{tikzpicture}
		\begin{axis}[
			width=\scatterplotsize,
			height=\scatterplotsize,
			axis equal image,
			xmin=0,
			ymin=0,
			ymax=4000,
			xmax=4000,
			xmode=log,
			ymode=log,
			axis x line=bottom,
			axis y line=left,
			extra x ticks = {4000},
			extra x tick labels = {TO},
			extra x tick style = {grid = major},
			extra y ticks = {4000},
			extra y tick labels = {TO},
			extra y tick style = {grid = major},
			xlabel={#3},
			xlabel style={font=\scriptsize,yshift=5pt},
			ylabel={#5},
			ylabel style={font=\scriptsize,yshift=-9pt},
			yticklabel style={font=\scriptsize},
			xticklabel style={font=\scriptsize},
			legend pos=north east,
			legend columns=5,
			legend style={nodes={scale=0.75, transform shape},inner sep=1.5pt, xshift=1mm, yshift=7mm},
			set layers,
			mark layer=axis background
			]
			
			\iftoggle{showplots}{\addplot[
				scatter,
				only marks,
				scatter/classes={
					cutoffstrategy={mark=square*,color1,mark size=1.25},
					distribution={mark=triangle*,color2,mark size=1.25},
					minimizeautomaton={mark=o,color3,mark size=1.75}
				},
				scatter src=explicit symbolic
				]%
				table [col sep=tab,x=#2,y=#4,meta=heuristic
				] {#1};
			}{\node[anchor=south west, align=center, red] {\huge NOT\\ COMPILED};}
			\ifthenelse{\NOT\equal{#6}{false}}{\legend{base,H1,H2}}{}
			\addplot[no marks, name path=f] coordinates {(0.001,0.001) (4000,4000) };
			\addplot[name path=g] coordinates {(0.001,4000) (4000,4000)};
		\end{axis}
	\end{tikzpicture}
}

\newcommand{\scatterplotRmin}[6]{%
	\begin{tikzpicture}
		\begin{axis}[
			width=\scatterplotsize,
			height=\scatterplotsize,
			axis equal image,
			xmin=0,
			ymin=0,
			ymax=20000,
			xmax=20000,
			xmode=log,
			ymode=log,
			axis x line=bottom,
			axis y line=left,
			extra x ticks = {20000},
			extra x tick labels = {TO},
			extra x tick style = {grid = major},
			extra y ticks = {20000},
			extra y tick labels = {TO},
			extra y tick style = {grid = major},
			xlabel={#3},
			xlabel style={font=\scriptsize,yshift=5pt},
			ylabel={#5},
			ylabel style={font=\scriptsize,yshift=-9pt},
			yticklabel style={font=\scriptsize},
			xticklabel style={font=\scriptsize},
			legend pos=north east,
			legend columns=5,
			legend style={nodes={scale=0.75, transform shape},inner sep=1.5pt, xshift=1mm, yshift=7mm},
			set layers,
			mark layer=axis background
			]
			\iftoggle{showplots}{\addplot[
				scatter,
				only marks,
				scatter/classes={
					cutoffstrategy={mark=square*,color1,mark size=1.25},
					distribution={mark=triangle*,color2,mark size=1.25},
					minimizeautomaton={mark=o,color3,mark size=1.75}
				},
				scatter src=explicit symbolic
				]%
				table [col sep=tab,x=#2,y=#4,meta=heuristic
				] {#1};
			}{\node[anchor=south west, align=center, red] {\huge NOT\\ COMPILED};}
			\ifthenelse{\NOT\equal{#6}{false}}{\legend{base,H1,H2}}{}
			\addplot[no marks, name path=f] coordinates {(0.001,0.001) (20000,20000) };
			\addplot[name path=g] coordinates {(0.001,0.001) (20000,0.001)};
		\end{axis}
	\end{tikzpicture}
}

\newcommand*\cir[1]{\tikz[baseline=(char.base),font=\small\bfseries]{%
		\node[shape=circle,draw,inner sep=0.5pt] (char) {{#1}};}}

\newcommand*\circled[1]{\kern-2.5em%
	\put(0,3){\color{white}\circle*{11}}\put(0,3){\circle{9}}%
	\put(-2.8,0){\color{black}\bfseries\small#1}~~}

\tikzstyle{new style 0}=[fill=none, draw=none, shape=circle]
\tikzstyle{none}=[fill=none, draw=none, shape=circle]
\tikzstyle{left-align}=[fill=none, shape=circle, align=left]
\tikzstyle{rectangle}=[fill={rgb,255: red,232; green,232; blue,232}, draw=black, shape=rectangle, minimum width=4cm, minimum height=1cm]
\tikzstyle{center-align}=[fill=none, draw=none, shape=circle, align=center]
\tikzstyle{rectangle-white}=[fill=white, draw=black, shape=rectangle, align=center, minimum width=2.2cm, minimum height=1.5cm]
\tikzstyle{circle-center}=[fill=none, draw=black, shape=circle, align=center, minimum size=1.5cm]
\tikzstyle{tiny}=[fill=black, draw=black, shape=circle, minimum size=3pt, inner sep=0pt, outer sep=0pt]
\tikzstyle{circle1}=[fill=none, draw=black, shape=circle, minimum size=1.2cm]
\tikzstyle{rect-teal}=[fill={rgb,255: red,214; green,218; blue,247}, draw=black, shape=rectangle, minimum width=2.2cm, minimum height=1.5cm, align=center, rounded corners=0.5cm]
\tikzstyle{rect-lime}=[fill={rgb,255: red,229; green,255; blue,135}, draw=black, shape=diamond, minimum width=2.2cm, minimum height=1.5cm, align=center, rounded corners=0.5cm]
\tikzstyle{rect-pink}=[fill={rgb,255: red,255; green,211; blue,224}, draw=black, shape=rectangle, minimum width=2.2cm, minimum height=1.5cm, align=center, rounded corners=0.5cm]
\tikzstyle{center-align}=[fill=none, draw=none, shape=circle, align=center]
\tikzstyle{arrow 2}=[dash pattern=on 3pt off 3pt]
\tikzstyle{right-align}=[fill=none, draw=none, shape=circle, align=right]
\tikzstyle{rect-yellow}=[fill={rgb,255: red,255; green,251; blue,189}, draw=black, shape=rectangle, minimum width=2.2cm, minimum height=1.5cm, align=center, rounded corners=0.5cm]

\tikzstyle{arrow}=[->]
\tikzstyle{line}=[-, fill=white]
\tikzstyle{green-box-line}=[-, fill={rgb,255: red,197; green,245; blue,186}]
\tikzstyle{thin line}=[-, line width=1pt]
\tikzstyle{yellow-box-line}=[-, fill={rgb,255: red,247; green,255; blue,170}]
\tikzstyle{new edge style 0}=[->]
\tikzstyle{gray box}=[-, fill={rgb,255: red,244; green,244; blue,244}]
\tikzstyle{white-box}=[-, fill=white]
\tikzstyle{dotted}=[-, dotted]
\tikzstyle{blue box}=[-, fill={rgb,255: red,208; green,241; blue,245}]

\definecolor{orcidlogocol}{HTML}{A6CE39}
\def\orcidID#1{\unskip$^{\smash{\href{http://orcid.org/#1}{\protect\raisebox{-1.25pt}{\protect\includegraphics{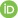}}}}}$}

\title{Learning Explainable and Better Performing Representations of POMDP Strategies
\thanks{
	This research was funded in part by the German Research Foundation (DFG) project 427755713 GOPro, the MUNI Award in Science and Humanities (MUNI/I/1757/2021) of the Grant Agency of Masaryk University, the DFG research training group GRK 2428 Continuous Verification of Cyber-Physical Systems (ConVeY) and the DFG RTG 2236/2
	(UnRAVeL).}
}

\author{}\institute{}
\author{
	Alexander Bork\inst{1}\orcidID{0000-0002-7026-228X} \and
	Debraj Chakraborty\inst{2}\orcidID{0000-0003-0978-4457} \and
	Kush Grover\inst{3}\orcidID{0000-0003-4575-1302} \and
	Jan K\v{r}et\'insk\'y\inst{2,3}\orcidID{0000-0002-8122-2881} \and
	Stefanie Mohr\inst{3}\orcidID{0000-0002-8630-3218}\textsuperscript{(\Letter)}
}

\authorrunning{A. Bork et al.}

\institute{RWTH Aachen University, Aachen, Germany \\
	\email{firstname.lastname@cs.rwth-aachen.de} \\ \and
	Masaryk University, Brno, Czechia  \\
	\email{lastname@fi.muni.cz}\\ \and
	Technical University of Munich, Munich, Germany\\
	\email{firstname.lastname@tum.de}}

\begin{document}

\maketitle

\begin{abstract}
Strategies for partially observable Markov decision processes (POMDP) typically require memory.
One way to represent this memory is via automata. 
We present a method to learn an automaton representation of a strategy using a modification of the $L^*$-algorithm.
Compared to the tabular representation of a strategy, the resulting automaton is dramatically smaller and thus also more explainable.
Moreover, in the learning process, our heuristics may even improve the strategy's performance.
In contrast to approaches that synthesize an automaton directly from the POMDP thereby solving it, our approach is incomparably more scalable.
\end{abstract} 

\section{Introduction}

\paragraph{Partially Observable Markov Decision Processes (POMDPs)} combine non-determinism, probability and partial observability.
Consequently, they have gained popularity in various applications as a model of planning in an unsafe and only partially observable environment. 
Coming from the machine learning community \cite{russell2010artificial}, they also gained interest in the formal methods community \cite{madani2003undecidability,bork2022under,chatterjee2016decidable,KAELBLING199899}.
They are a very powerful model, able to faithfully capture real-life scenarios where we cannot assume perfect knowledge, which is very often the case.
Unfortunately, the great power comes with the hardness of analysis.
Typical objectives of interest such as reachability or total reward already result in undecidable problems \cite{madani2003undecidability}.
Namely, the resolution of the non-determinism (a.k.a.\ synthesis of a \emph{strategy}, policy, scheduler, or controller) cannot be done algorithmically while guaranteeing optimality w.r.t.\ the objective.
Consequently, \emph{heuristics} to synthesize practically well-performing strategies became of significant interest.
Let us name several aspects playing a key role in applicability of such synthesis procedures:
\begin{enumerate}[label=\protect\circled{\arabic*}]
	\item \emph{quality} of the synthesized strategies,
	\item \emph{size} and \emph{explainability} of the representation of the synthesized strategies,
	\item \emph{scalability} of the computation method.
\end{enumerate}

\paragraph{Strategy Representation.}
While \cir{1} and \cir{3} are of obvious importance, it is important to note the aspect \cir{2}.
A strategy is a function mapping the current history (sequence of observations so far) to an action available in the current state.
When written as a list of history-action pairs, it results in a large and incomprehensible table.
In contrast, when equivalently written as a Mealy machine transducing the stream of observation to a stream of actions,
its size may be dramatically lower (making it easier to implement and more efficient to execute) and its representation more explainable (making it easier to certify).
Besides, better understandability allows for easier maintenance and modification.
To put it in a contrast, explicit (table-like) or, e.g., neural-network representations of the function can hardly be hoped to be understandable by any human (even domain expert).
Compact and understandable representations of strategies have recently gained attention, e.g., \cite{DBLP:conf/cav/BrazdilCCFK15,DBLP:conf/tacas/NeiderT16}, also for POMDP \cite{junges2018,andriushchenko2022}, and even tool support \cite{DBLP:conf/tacas/AshokJKWWY21} and \cite{paynt}, respectively.
See \cite{DBLP:conf/hybrid/AshokJJKWZ20} for detailed aspects of motivation for compact representations.

\paragraph{Current Approaches}
For POMDP, the state of the art is torn into two streams.

On the one hand, tools such as \tool{Storm} \cite{bork2022under} feature a classic \emph{belief-based} analysis, which essentially blows up the state space, making it easier to analyze.
Consequently, it is still reasonably scalable \cir3, but the size of the resulting strategy is even larger than that of the state space of the POMDP and is simply given as a table, i.e., not doing well w.r.t. the representation \cir2. 
Moreover, to achieve the scalability (and in fact even termination), the analysis has to be stopped at some places (``cut-offs''), resulting in poorer performance \cir1.
On the other hand, the exhaustive bounded synthesis as in \tool{PAYNT} \cite{paynt} tries to synthesize a small Mealy machine representing a good strategy (while thus solving the POMDP) and if it fails, it tries again with an increased allowed size of the automaton.
While this approach typically achieves better quality \cir1 and, by principle, better size and explainability \cir2, it is extremely expensive and does not scale at all if the strategy requires a larger automaton \cir3.
While symbiotic approaches are emerging \cite{andriushchenko2023}, the best of both worlds has not been achieved yet.

\paragraph{Our Contribution}
We design a highly scalable postprocessing step, which improves the quality and the representation of the strategy.
It is compatible with any framework producing any strategy representation, requiring only that we can query the strategy function (which action corresponds to a given observation sequence).
In particular, \tool{Storm}, which itself is scalable, can thus profit from improving the quality and the representation of the produced strategies.
Our procedure learns a compact representation of the given strategy as a Mealy machine using \emph{automata-learning} techniques, in two different ways.
First, through learning the complete strategy, we get its automaton representation, which is \emph{fully equivalent} and thus achieving also the same value.
Second, we provide \emph{heuristics} learning small modifications of the strategy.
Indeed, for some inputs (observation sequences), we ignore what the strategy suggests, in particular when the strategy is not defined, but also when it explicitly states that it is unsure about its choice (such as at the cut-off points, where the sequences become too long and the strategy was not optimised well at these later points).
Whenever we ignore the strategy, we try to devise with a possibly better solution.
For instance, we can adopt the decision that the currently learnt automaton suggests, or we can reflect other decisions in similar situations.
This way we produce a \emph{simpler strategy} (thus also comparatively smaller), which can, in principle, \emph{fix the suboptimal decisions} of the strategy stemming from the limitations of the original analysis (such as bounds on the exploration) or any other irregularities.
Of course, this only works well if the true optimal strategy is ``sensible'', i.e., has inner structure allowing for a simple automaton representation.
For practical, hence sensible, problems, this is typically the case.

\noindent \emph{Summary of our contribution:}
\begin{itemize}
	\item We provide a method to take any POMDP strategy and transform it into an equivalent or similar (upon choice) automaton, yielding \textbf{small} size and potential for \textbf{explainability}.
	\item Thereby we often improve the \textbf{quality} of the strategy.
	\item The experiments confirm the improvements and frequent proximity to best known values (typically of PAYNT) on the simpler benchmarks.
	\item The experiments confirm great \textbf{scalability} even on harder benchmarks, which are out of reach of PAYNT. 
	Hence no comparison in quality or size is really possible here. 
	However, the auspicious comparison on simpler ones warrants the trust in good absolute quality and size on the harder ones (and the results are anyway at least the best among the available state-of-the-art).
\end{itemize}

\paragraph{Related Work}
Methods to solve planning problems on POMDPs have been studied extensively in the literature \cite{smallwood1973,hauskrecht2000,shani2013survey}.
Many state-of-the-art solvers use point-based methods like \emph{PBVI} \cite{pineau2003}, \emph{Perseus} \cite{spaan2005} and \emph{SARSOP} \cite{kurniawati2008} to treat bounded and unbounded discounted properties.
For these methods, strategies are typically represented using so called $\alpha$-vectors. 
Apart from a significant overhead in the analysis, they completely lack of explainability.
Notably, while the \emph{SARSOP} implementation provides an export of its computed strategies in an automaton format, we have not been able to find an explanation of how it is generated.

Methods based on the (partial) exploration and solving of the belief MDP underlying the POMDP \cite{norman2017,bork2020,bork2022under} have been implemented in the probabilistic model checkers \tool{Storm} \cite{storm} and \tool{Prism} \cite{prism}.
The focus of these methods is optimizing infinite-horizon objectives \emph{without} discounting.
Recent work \cite{andriushchenko2023} describes how strategies are extracted from the results of these belief exploration methods. 
The resulting strategy representation, however, is rather large and contains potentially redundant information.

Orthogonal to the methods above, there are approaches that \emph{directly} synthesize strategies from a space of candidates \cite{hansen1998,meuleau1999}. 
The synthesized strategy is then applied to the POMDP to yield a Markov chain. Analyzing this Markov chain yields the objective value achieved by the strategy.
Methods used for searching policies include using inductive synthesis \cite{andriushchenko2022}, gradient decent \cite{heck2022} or convex optimization \cite{amato2010,junges2018,cubuktepe2021}.
\cite{andriushchenko2023} describes an integration of a belief exploration approach \cite{bork2022under} with inductive synthesis \cite{andriushchenko2022}.

Our approach is orthogonal to the solution methods in that it uses an \emph{existing} strategy representation and learns a new, potentially more concise finite-state controller representation.
Furthermore, our modifications of learned strategy representations shares similarities with approaches for strategy improvement \cite{thomas2015,carr2019,aaai23safe}.

\section{Preliminaries}
For a countable set $S$, we denote its power set by $2^S$. A \emph{(discrete) probability distribution} on a countable set $S$ is a function $d:S\to [0,1]$ such that $\sum_{s\in S} d(S)=1$.
We denote the set of all probability distributions on the set $S$ as $\distribution(S)$. 
For $d \in \distribution(S)$, the \emph{support} of $d$ is $\supp(d) = \{s \in S \mid d(s) > 0\}$. 
We use the Iverson bracket notation where $[x]=1$ if the expression $x$ is true and $0$ otherwise. 
For two sets $S,T$, we define the set of concatenations of $S$ with $T$ as $S \cdot T = \{s \cdot t \mid s \in S, t \in T\}$. We analogously define the set of $n$-times concatenation of $S$ with itself as $S^n$ for $n \geq 1$ and $S^0 =\{\epsilon\}$ is the set containing the empty string.
We denote by $S^* = \bigcup_{i=0}^\infty S^n$ the set of all \emph{finite strings} over $S$ and by $S^+ = \bigcup_{i=1}^\infty S^n$ the set of all \emph{non-empty} finite strings over $S$.
For a finite string $w=w_1w_2\ldots w_n$, the string $w[0,i]$ with $w[0,0] = \epsilon$ and $w[0,i] = w_1\ldots w_i$ for $0 < i \leq n$ is a \emph{prefix} of $w$.
The string $w[i,n] = w_i\ldots w_n$ with $0 < i \leq n$ is a \emph{suffix} of $w$.
A set $W \subseteq S^*$ is \emph{prefix-closed} if for all $w \in S^*$, $w = w_1 \ldots w_n \in W$ implies $w[0, i] \in W$ for all $0 \leq i \leq n$. 
A set $W' \subseteq S^*$ is \emph{suffix-closed} if $\epsilon \notin W$ and for all $w \in S^*$, $w = w_1 \ldots w_n \in W$ implies $w[i, n] \in W$ for all $0 < i \leq n$.

\begin{definition}[MDP]
    A \emph{Markov decision process} (MDP) is a tuple $\mdp = (\mdpstates, \mdpactions, \mdptransitions, \mdpinitstates)$ where $\mdpstates$ is a countable set of states, $\mdpactions$ is a finite set of actions, 
    $\mdptransitions: \mdpstates\times \mdpactions\rightharpoonup \distribution(\mdpstates)$ is a partial transition function,
    and $\mdpinitstates\in\mdpstates$ is the initial state. 
\end{definition}
For an MDP $\mdp = (\mdpstates, \mdpactions, \mdptransitions, \mdpinitstates)$, $s\in \mdpstates$ and $a\in \mdpactions$, let $\post[\mdp](s,a)=\{s'\mid \mdptransitions(s,a,s')>0\}$ be the set of successor states of $s$ in $\mdp$ that can be reached by taking the action $a$. 
We also define the set of \emph{enabled actions} in $s\in \mdpstates$ by $\mdpavailactions(s) = \{a\in \mdpactions \mid \mdptransitions(s,a)\neq\bot\}$.
A \emph{Markov chain} (MC) is an MDP with $|\mdpavailactions(s)|=1$ for all $s\in\mdpstates$. 
For an MDP $\mdp$, a \emph{finite path} $\mdppath = s_0a_0s_1\ldots s_i$ of length $i\ge 0$ is
a sequence of states and actions such that for all $t\in[0,i-1]$, $a_t\in \mdpavailactions(s_t)$ and $s_{t+1}\in \post[\mdp](s_t,a_t)$.
Similarly, an infinite path is an infinite sequence $\mdppath = s_0a_0s_1a_1s_2\ldots$ such that 
for all $t\in\N$, $a_t\in \mdpavailactions(s_t)$ 
and $s_{t+1}\in \post[\mdp](s_t,a_t)$.
For an MDP $\mdp$, we denote the set of all finite paths by $\mdpfinitepaths$, and of all infinite paths by $\mdpinfinitepaths$.
\begin{definition}[POMDP]
    A \emph{partially observable MDP (POMDP)} is a tuple $\pomdp =(\mdp, \observations, \observationmap)$ where $\mdp=(\mdpstates, \mdpactions, \mdptransitions, \mdpinitstates)$ is the underlying MDP with finite number of states, $\observations$ is a finite set of observations, and $\observationmap:\mdpstates\rightarrow\observations$ is an observation function that maps each state to an observation.
\end{definition}

For POMDPs, we require that states with the same observation have the same set of enabled actions, i.e., $\observationmap(s) = \observationmap(s')$ implies $\mdpavailactions(s) = \mdpavailactions(s')$ for all $s,s'\in S$. This way, we can lift the notion of enabled actions to an observation $z\in\observations$ by setting $\mdpavailactions(z)=\mdpavailactions(s)$ for some state $s\in \mdpstates$ with $\observationmap(s)=z$. 
The notion of observation $\observationmap$ for states can be lifted to paths: for a path $\mdppath = s_0 a_0 s_1 a_1 \ldots$, we define $\observationmap(\mdppath) = \observationmap(s_0)a_0\observationmap(s_1)a_1\ldots$. Two paths $\rho_1$ and $\rho_2$ are called \emph{observation-equivalent} if $\observationmap(\rho_1) = \observationmap(\rho_2)$.
We call an element $\observationsequence \in \observations^*$ an \emph{observation sequence} and denote the observation sequence of a path $\mdppath = s_0a_0s_1\ldots$ by $\observationsequenceof(\mdppath) = \observationmap(s_0)\observationmap(s_1)\ldots$ .
\begin{wrapfigure}[10]{l}{0.25\textwidth}
	\vspace{-29pt}
	\vspace{20pt}
	\centering
	\resizebox{0.2\textwidth}{!}{\begin{tikzpicture}
	\begin{pgfonlayer}{nodelayer}
		\node [style=none] (0) at (0, 0) {};
		\node [style=none] (1) at (2, 0) {};
		\node [style=none] (2) at (-2, 2) {};
		\node [style=none] (3) at (-2, 0) {};
		\node [style=none] (4) at (2, -2) {};
		\node [style=none] (5) at (0, -2) {};
		\node [style=none] (8) at (-2, 0) {};
		\node [style=none] (11) at (0, 2) {};
		\node [style=none] (12) at (0, 0) {};
		\node [style=none] (13) at (0, 0) {};
		\node [style=none] (14) at (0, -2) {};
		\node [style=none] (15) at (-2, -2) {};
		\node [style=none] (16) at (0, 2) {};
		\node [style=none] (17) at (2, 2) {};
		\node [style=none] (18) at (2, 0) {};
		\node [style=none] (19) at (0, 0) {};
		\node [style=center-align] (20) at (-1, 1) {\huge 0};
		\node [style=center-align] (20) at (-1.7, 1.7) {\Large\texttt{b}};
		\node [style=center-align] (21) at (1, 1) {\huge 1};
		\node [style=center-align] (20) at (0.3, 1.7) {\Large\texttt{y}};
		\node [style=center-align] (22) at (-1, -1) {\huge 2};
		\node [style=center-align] (20) at (-1.7, -0.3) {\Large\texttt{b}};
		\node [style=center-align] (23) at (1, -1) {\huge 3};
		\node [style=center-align] (20) at (0.3, -0.3) {\Large\texttt{g}};
	\end{pgfonlayer}
	\begin{pgfonlayer}{edgelayer}
		\draw [style=green-box-line] (4.center)
			 to (5.center)
			 to (0.center)
			 to (1.center)
			 to cycle;
		\draw [style=blue box] (2.center)
			 to (3.center)
			 to (12.center)
			 to (11.center)
			 to cycle;
		\draw [style=blue box] (13.center)
			 to (14.center)
			 to (15.center)
			 to (8.center)
			 to cycle;
		\draw [style=yellow-box-line] (19.center)
			 to (16.center)
			 to (17.center)
			 to (18.center)
			 to cycle;
	\end{pgfonlayer}
\end{tikzpicture}}
	\caption{Running\\ example: POMDP}
	\label{fig:running-example}
\end{wrapfigure}
\begin{example}\label{ex:pomdp}
	Consider the POMDP graphically depicted in \cref{fig:running-example}, modeling a basic robot planning task. 
	A robot is dropped uniformly at random in one of four grid cells. Its goal is to reach cell 3.
	The robot's sensors cannot to distinguish cells 0 and 2, while cells 1 and 3 provide unique information.
	For the POMDP model, we use states $0$, $1$, $2$, and $3$ to indicate the robot's position. 
	We mimic the random initialization by introducing a unique initial state $s_0$ with a unique observation $\texttt{i}$ (init). 
	$s_0$ has a single action that reaches any of the other four states with equal probability $0.25$. 
	Thus, the state space of the POMDP is $\mdpstates=\{s_0,0,1,2,3\}$. To represent the observations of the robot, we use three observations $\texttt{b}$, $\texttt{y}$ and $\texttt{g}$, so $\observations=\{\texttt{i},\texttt{b},\texttt{y},\texttt{g}\}$. 
	States $0$ and $2$ have the same observation, while states $1$ and $3$ are uniquely identifiable, formally $\observationmap=\{(s_0\rightarrow\texttt{i}), (0\rightarrow \texttt{b}),(1\rightarrow \texttt{y}), (2\rightarrow \texttt{b}), (3\rightarrow \texttt{g})\}$.
	The goal is for the robot to reach state $3$.
	In each state, it can choose to move up, down, left, or right, $\mdpactions=\{s,u,d,l,r\}$.
	In each step, executing the chosen action may fail with a probability of $p=0.5$, causing the robot to remain in its current cell without changing states.
\end{example}

\begin{definition}[Strategy]
A strategy for an MDP $\mdp$ is a funtion $\mdpstrategy : \mdpfinitepaths\to\distribution(\mdpactions)$ such that for all paths $\mdppath\in \mdpfinitepaths$, $\supp(\mdpstrategy(\mdppath))\subseteq\mdpavailactions(\last(\mdppath))$. 
\end{definition}

A strategy $\mdpstrategy$ is \emph{deterministic} if $|\supp(\mdpstrategy(\mdppath))| = 1$ for all paths $\mdppath\in \mdpfinitepaths$. Otherwise, it is \emph{randomized}. A strategy $\mdpstrategy$ is called \emph{memoryless} if it depends only on $last(\mdppath)$ \textit{i.e.} for any two paths $\mdppath_1,\mdppath_2 \in \mdpfinitepaths$, if $last(\mdppath_1) = last(\mdppath_2)$ then $\mdpstrategy(\mdppath_1) = \mdpstrategy(\mdppath_2)$.
As general strategies have access to \emph{full} state information, they are unsuitable for partially observable domains. Therefore, POMDPs require a notion of strategies based only on observations.
For a POMDP $\pomdp$, we call a strategy \emph{observation-based} if for any $\mdppath_1,\mdppath_2 \in \mdpfinitepaths$, $\observationmap(\mdppath_1) = \observationmap(\mdppath_2)$ implies $\mdpstrategy(\mdppath_1) = \mdpstrategy(\mdppath_2)$, i.e.\ the strategy has same output on observation-equivalent paths.

We are interested in representing observation-based strategies approximating optimal objective values for infinite horizon objectives without discounting, also called \emph{indefinite-horizon objectives}, i.e.,\ maximum/minimum reachability probabilities and expected total reward objectives.
We emphasize that our general learning framework also generalizes straightforwardly to strategies for different objectives.
In contrast to fully observable MDPs, deciding if a given strategy is optimal for an indefinite-horizon objective on a POMDP is generally undecidable \cite{madani2003undecidability}.
In fact, optimal behavior requires access to the \emph{full} history of observations, necessitating an arbitrary amount of memory.
As such, our goal is to learn a small representation of a strategy using only a finite amount of memory that approximates optimal values as well as possible.

We represent these strategies as \emph{finite-state controllers (FSCs)} -- automata that compactly encode strategies with access to memory and randomization in a POMDP.
\begin{definition}[Finite-State Controller]
A \emph{finite-state controller} (FSC) is a tuple $\fsc = (\fscstates, \fscactionmap, \fsctransitionmap, \fscinitstate)$ where $\fscstates$ is a finite set of nodes, $\fscactionmap: \fscstates \times \observations \rightarrow \distribution(\mdpactions)$ is an action mapping, $\fsctransitionmap:\fscstates \times \observations \rightarrow \fscstates$ is the transition function, and $\fscinitstate$ is the initial node.
\end{definition}
We denote by $\mdpstrategy_{\fsc}$ the strategy represented by the FSC $\fsc$ and use $\fscset$ for the set of all FSCs for a POMDP $\pomdp$.
Given an FSC $\fsc = (\fscstates, \fscactionmap, \fsctransitionmap, \fscinitstate)$ that is currently in node $n$, and a POMDP $\pomdp$ with underlying MDP $\mdp = (\mdpstates, \mdpactions, \mdptransitions, \mdpinitstates)$, in state $s$, the action to play by an agent following $\fsc$ is chosen randomly from the distribution $\fscactionmap(n,\observationmap(s))$. 
$\fsc$ then updates its current node to $n' = \fsctransitionmap(n,z)$. The state of the POMDP is updated according to $\mdptransitions$.
As such, an FSC induces a Markov chain $\mdp_{\fsc} = (\mdpstates \times \fscstates, \{\alpha\}, \mdptransitions^{\fsc}, (s_0,n_0))$ where
$\mdptransitions^{\fsc}((s,n),\alpha,(s',n'))$ is $[\fsctransitionmap(n,\observationmap(s)) = n'] \cdot
\sum_{a\in\mdpavailactions(s)}\fscactionmap(n,\observationmap(s))(a) \cdot \mdptransitions(s,a,s')$.

An FSC can be interpreted as a Mealy machine: nodes correspond directly to states of the Mealy machine, which takes observations as input.
The set of output symbols is the set of all distributions over actions occurring in the FSC.

\section{Learning a Finite-State Controller}\label{sec:learning-fsc}
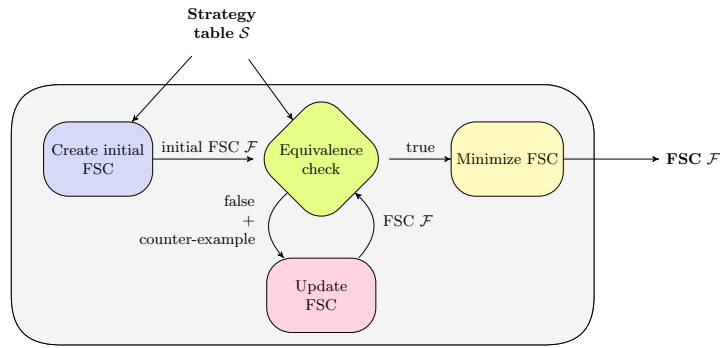
\begin{figure}[t]
	\centering
	\resizebox{0.8\textwidth}{!}{
		\begin{tikzpicture}
	\begin{pgfonlayer}{nodelayer}
		\node [style=none] (14) at (-3.75, 2.25) {};
		\node [style=none] (15) at (-2.75, 3.25) {};
		\node [style=none] (16) at (7, 3.25) {};
		\node [style=none] (17) at (8, 2.25) {};
		\node [style=none] (18) at (8, -1) {};
		\node [style=none] (19) at (7, -2) {};
		\node [style=none] (20) at (-2.75, -2) {};
		\node [style=none] (21) at (-3.75, -1) {};
		\node [style=rect-teal] (1) at (-2, 1.75) {Create initial\\FSC};
		\node [style=rect-lime] (3) at (2.5, 1.75) {Equivalence\\check};
		\node [style=rect-pink] (5) at (2.5, -1) {Update\\FSC};
		\node [style=center-align] (7) at (0.5, 4.5) {\textbf{Strategy}\\\textbf{table} \strategytable};
		\node [style=none] (9) at (4.25, 0.5) {FSC $\fsc$};
		\node [style=none] (10) at (4.5, 2) {true};
		\node [style=none] (12) at (0.25, 2) {initial FSC $\fsc$};
		\node [style=center-align] (13) at (10, 1.75) {\textbf{FSC} $\fsc$};
		\node [style=right-align] (24) at (0, 0.5) {false\\+\\counter-example};
		\node [style=rect-yellow] (25) at (6.25, 1.75) {Minimize FSC};
	\end{pgfonlayer}
	\begin{pgfonlayer}{edgelayer}
		\draw [style=gray box] (19.center)
			 to (20.center)
			 to [bend left=45, looseness=1.25] (21.center)
			 to (14.center)
			 to [bend left=45, looseness=1.25] (15.center)
			 to (16.center)
			 to [bend left=45] (17.center)
			 to (18.center)
			 to [bend left=45, looseness=1.25] cycle;
		\draw [style=gray box] (16.center)
			 to [bend left=45] (17.center)
			 to (18.center)
			 to [bend left=45, looseness=1.25] (19.center)
			 to (20.center)
			 to [bend left=45, looseness=1.25] (21.center)
			 to (14.center)
			 to [bend left=45, looseness=1.25] (15.center)
			 to cycle;
		\draw [style=arrow] (7) to (1);
		\draw [style=arrow] (7) to (3);
		\draw [style=arrow] (1) to (3);
		\draw [style=arrow, bend right=45, looseness=1.25] (3) to (5);
		\draw [style=arrow, bend right=45, looseness=1.25] (5) to (3);
		\draw [style=arrow] (3) to (25);
		\draw [style=arrow] (25) to (13);
	\end{pgfonlayer}
\end{tikzpicture}}
	\caption{Depiction of the FSC learning framework}
	\label{fig:flow-chart}
\end{figure}
We present a framework for learning a concise finite-state controller representation from a given strategy for a POMDP.
Our approach mimics an extension of the L* automaton learning approach \cite{angluin1987learning} for learning Mealy machines \cite{mealymachines}. 
The main difference in our approach is that we have a sparse learning space: not all observations of a POMDP are possible to reach from all states.
Thus, there are many observation sequences that can never occur in the POMDP.
To mark situations where this occurs, i.e. where a learned FSC has complete freedom to decide what to do, we introduce a ``don't-care" symbol $\dontcare$.

Furthermore, for some policy computation methods, the strategy we receive as input may be incomplete. 
Although some observation sequence can appear in the POMDP, 
the strategy does not specify what to do when it occurs. 
This can for example be caused by reaching the depth limit in an exploration based approach. We use a ``don't-know" symbol $\dontknow$ to mark such cases.
While the non-occuring sequences do not directly influence the learning process, they cannot be ignored completely. 
These $\dontknow$ need to be replaced by actual actions using some heuristics for the final FSC to yield a complete strategy (see \Cref{sec:cutoffHeuristics}).

An overview of the learning process is depicted in \cref{fig:flow-chart}.
We expect as input a (partially defined) strategy in the form of a table that maps observation sequences in the POMDP to a distribution over actions. 
\begin{definition}[Strategy Table]
	A \emph{strategy table} $\strategytable$ for a POMDP $\pomdp$ is a relation $\strategytable \subseteq \observations^* \times (\distribution(\mdpactions) \cup \{\dontknow\})$.
	A \emph{row} of $\strategytable$ is an element $(\observationsequence,d) \in \strategytable$.
\end{definition}

For $(\observationsequence,d) \in \strategytable$, if $\supp(d)$ contains only a single action $a$, we write it as $(\observationsequence,a)$.
We say a strategy table $\strategytable$ is \emph{consistent} if and only if for $\observationsequence \in \observations^*$, $(\observationsequence,d_1) \in \strategytable$ and $(\observationsequence,d_2) \in \strategytable$ implies $d_1 = d_2$, i.e. each observation sequence has at most one unique output.
A consistent strategy table $\strategytable$ (partially) defines an observation-based strategy $\mdpstrategy_{\strategytable}$ with $\mdpstrategy_{\strategytable}(\mdppath) = d$ if and only if $(\observationsequenceof(\mdppath), d) \in \strategytable$ and $d \neq \dontknow$.
For consistent strategy tables, the FSC resulting from our approach correctly represents the partially defined strategy.

\begin{example}
\cref{tab:lookup-table} depicts a strategy table for the POMDP described in \cref{ex:pomdp}. 
The table does not specify what to do in state $3$ as at that point, the robot has already achieved its target.
The action chosen at that point is irrelevant.
Intuitively, the strategy table describes that the robot should go right as long as it sees \texttt{b}, and goes down once it sees \texttt{y}.
The FSC in \Cref{fig:strategy-fsc} fully captures the behaviour described by the strategy table and thus accurately represents it.
\end{example}

\begin{figure}[t]
	\begin{minipage}{0.6\textwidth}
		\centering
		\scalebox{0.9}{
			\begin{tabular}{c|c}
				Observation sequence & Action\\
				\midrule
				\texttt{i} & s \\
				\texttt{i y} & d \\
				\texttt{i b} & r \\
		\end{tabular}}
		\captionof{table}{Example strategy table for the POMDP in \cref{ex:pomdp}. It only contains observation sequences of length at most 2.}
		\label{tab:lookup-table}
	\end{minipage}
	\hfill	
	\begin{minipage}{0.35\textwidth}
		\centering
		\resizebox{0.5\textwidth}{!}{
			\begin{tikzpicture}
	\node[style=state] (1) {};
	\node[left=of 1, xshift=2.5cm] (0) {};
	\draw[] (0) edge[above] node{} (1);
	\draw[] (1) edge[loop above] node{\texttt{b}: r} (1);
	\draw[] (1) edge[loop right] node{\texttt{y}: d} (1);
	\draw[] (1) edge[loop below] node{\texttt{i}: s} (1);
\end{tikzpicture}}
		
		\caption{FSC representing the strategy table of \cref{tab:lookup-table}.}
		\label{fig:strategy-fsc}
	\end{minipage}
\end{figure}

In our framework, the input strategy table is used to build an initial FSC which is then compared to the input. 
If the initial FSC is already equivalent to the given strategy table, we are done and we output the FSC.
Otherwise, we get a counterexample and use it to update the FSC. 
This process of checking for equivalence and updating the FSC is repeated until the FSC is equivalent to the table.

In the sequel, we first explain how our learning approach works on general input of the form described above. 
Then we show how the learning approach is integrated with an existing POMDP solution method by means of the belief exploration framework from \cite{bork2022under}.
Lastly, we introduce heuristics for improvement of the learned policies when the information in the table is incomplete. 

\subsection{Automaton Learning}\label{sec:automaton-learning-basics}
The regular L* approach is used to learn a DFA for a regular language.
It is intuitively described as: a \emph{teacher} has information about an automaton and a \emph{student} wants to learn that automaton.
The student asks the teacher whether specific words are part of the language (\emph{membership query}).
At some point, the student proposes a solution candidate (in case of L*, a DFA) and asks the teacher whether it is correct, i.e. whether the proposed automaton accepts the language (\emph{equivalence query}).
Instead of the membership query of standard L*, the extension to Mealy machines \cite{mealymachines} uses an \emph{output query}, since we are not interested in the membership of a word in a language but rather the output of the Mealy machine corresponding to a specific word.
As such, our learning approach needs access to an output query, specifying the output of the strategy table for a given observation sequence, and an equivalence query, checking whether an FSC accurately represents the strategy table.
We formally define the two types of queries.

\begin{definition}[Output Query (OQ)]
The \emph{output query} for a strategy table $\strategytable$ is the function $OQ_{\strategytable}: \observations^* \to \distribution(\mdpactions) \cup \{\dontknow, \dontcare \}$ with 
$OQ_{\strategytable}(\observationsequence) = d$ if $(\observationsequence,d) \in \strategytable$ and $OQ_{\strategytable}(\observationsequence) = \dontcare$ otherwise.
\end{definition}

\begin{definition}[Equivalence Query (EQ)]
	The \emph{equivalence query} for a strategy table $\strategytable$ is a function $EQ_{\strategytable}: \fscset \to \observations^*$ defined as follows:
	$EQ_{\strategytable}(\fsc) = \epsilon$ if 
	for all $(\observationsequence, d) \in \strategytable$ and for all $\mdppath$ with $\observationsequenceof(\mdppath) = \observationsequence$,
	$\mdpstrategy_\fsc(\mdppath) = d $. 
	Otherwise, $EQ_{\strategytable}(\fsc) = c$ where	
	$c \in \{ \observationsequence \mid (\observationsequence, d) \in \strategytable, \exists \mdppath \in \mdpfinitepaths(\pomdp): \observationsequenceof(\mdppath) = \observationsequence \land \mdpstrategy_\fsc(\mdppath) \neq d\}$ is a \emph{counterexample} where $\strategytable$ and $\fsc$ have different output.
\end{definition}

The output query (OQ) takes an observation sequence $\observationsequence$, and outputs the distribution (or the $\dontknow$ symbol) suggested by the strategy table. If the given observation sequence is not present in the strategy table, it returns the $\dontcare$ symbol, i.e., a "don't care"-symbol. 
The equivalence query (EQ) takes a hypothesis FSC $\fschypothesis$ and asks whether it accurately represents $\strategytable$. 
In case it does not, an observation sequence where $\fschypothesis$ and $\strategytable$ differ is generated as a counterexample.

Using the definitions of these two queries, we formalise our problem statement as follows:\\
\noindent\framebox[\linewidth]{\parbox{0.95\linewidth}{
\textbf{Problem Statement:} Given a POMDP $\pomdp$, a strategy table $\strategytable$, an output query $OQ_{\strategytable}$ and an equivalence query $EQ_{\strategytable}$, compute a small FSC $\fsc$ such that $EQ_{\strategytable}(\fsc) = \epsilon$.
}}

\paragraph{Learning Table}\label{sec:learntable}
We aim at solving the problem using a learning framework similar to $L^*$.
We learn an FSC by creating a \emph{learning table} 
which keeps track of the observation sequences and the outputs the learner assumes they should yield in the strategy.
Formally, it is defined as follows:
\begin{definition}[Learning Table]
	A \emph{learning table} for POMDP $\pomdp$ is a tuple $\learningtable = ({\color{RawSienna} \upperrows}, {\color{OliveGreen}\columns}, {\color{Blue}\learntable})$ where
	${\color{RawSienna} \upperrows} \subset \observations^*$ is a prefix-closed finite set of finite strings over the observations representing the \emph{upper row indices}, the set ${\color{RawSienna} \upperrows} \cdot \observations$ are the \emph{lower rows indices} and ${\color{OliveGreen} \columns } \subset \observations^+$ is a suffix-closed finite set of non-empty finite strings over $\observations$ -- the \emph{columns}. 
	${\color{Blue}\learntable}:(\upperrows\cup\upperrows\cdot\observations)\times\columns\rightarrow\distribution(\mdpactions)\cup\{\dontknow,\dontcare\}$ is a mapping that represents the \emph{entries} of the table. 
\end{definition}

\begin{wrapfigure}[12]{l}{0.27\textwidth}
	\vspace{-20pt}
	\centering
	\begin{tabular}{b{10pt} | m{10pt}m{10pt}m{10pt}}
		& {\color{OliveGreen}\texttt{i}}&  {\color{OliveGreen}\texttt{b}} & {\color{OliveGreen}\texttt{y}} \\
		\toprule
		{\color{RawSienna}$\bm{\epsilon}$} & {\color{Blue}s}  & \color{Blue}$\dontcare$ & \color{Blue}$\dontcare$\\ 
		\midrule
		\texttt{i} & \color{Blue}$\dontcare$ & \color{Blue}r & \color{Blue}d \\
		\texttt{b} & \color{Blue}$\dontcare$ & \color{Blue}$\dontcare$ & \color{Blue}$\dontcare$\\
		\texttt{y} & \color{Blue}$\dontcare$ & \color{Blue}$\dontcare$ & \color{Blue}$\dontcare$\\
	\end{tabular}
	\makeatletter
	\def\@captype{table}
	\makeatother
	
	\caption{\\Running example - initial table}
	\label{tab:example-init-table}
\end{wrapfigure}

Intuitively speaking, the table is divided into {\color{RawSienna}\emph{upper}} and \emph{lower} rows.
Initially, the {\color{OliveGreen}columns} of the learning table are the observations in the POMDP.
Additional columns may be added in the learning process to further refine the behavior of the learned FSC.
Upper rows effectively result in nodes of the learned FSC, while lower rows specify destinations of the transitions.
For a row in the upper rows, each entry represents the output of the FSC corresponding to their respective observation (column).
For an upper row, if a column is labelled only with an observation, the corresponding {\color{Blue}entry} represents the output of the FSC on that observation.
As an example, \cref{tab:example-init-table} contains the initial learning table for our running example.
We do not include observation \texttt{g} for the target state as we are not interested in the behavior of the strategy after the target has been reached.

We say that two rows $r_1,r_2\in \upperrows\cup \upperrows\cdot\observations$ are \emph{equivalent} ($r_1\equiv r_2$) if they fully agree on their entries, i.e., $r_1\equiv r_2$ if and only if $\learntable(r_1,c) = \learntable(r_2,c)$ for all $c\in \columns$.
The equivalence class of a row $r \in \upperrows\cup \upperrows\cdot\observations$ is $[r] = \{ r' \mid r \equiv r' \}$.

\paragraph{From Learning Table to FSC}
To transform a learning table into an FSC, the table needs to be of a specific form. In particular, it needs to be \emph{closed} and \emph{consistent}.
A learning table is \emph{closed} if for each lower row $l \in \upperrows\cdot\observations$, there is an upper row $u\in\upperrows$ such that $l \equiv u$.

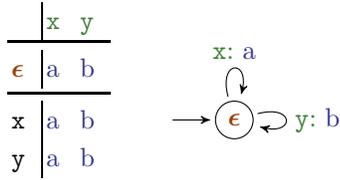
\begin{wrapfigure}[10]{l}{0.4\textwidth}
	\begin{minipage}{0.15\textwidth}
		\begin{tabular}{b{10pt} | m{10pt}m{10pt}}
			& {\color{OliveGreen}\texttt{x}}&  {\color{OliveGreen}\texttt{y}}\\
			\toprule
			{\color{RawSienna}$\bm{\epsilon}$} & {\color{Blue}a}  & \color{Blue}b\\ 
			\midrule
			\texttt{x} & {\color{Blue}a}  & \color{Blue}b\\ 
			\texttt{y} & {\color{Blue}a}  & \color{Blue}b\\ 
		\end{tabular}
	\end{minipage}\hfill
	\begin{minipage}{0.24\textwidth}
		\begin{tikzpicture}
	\node[style=state] (1) {\color{RawSienna}$\bm{\epsilon}$};
	\node[left=of 1, xshift=2.5cm] (0) {};
	\draw[] (1) edge[loop above] node{\color{OliveGreen}\texttt{x}: \color{Blue}a} (1);
	\draw[] (0) edge[above] node{} (1);
	\draw[] (1) edge[loop right] node{\color{OliveGreen}\texttt{y}: \color{Blue}b} (1);
\end{tikzpicture}
	\end{minipage}
	\caption{Transformation of a learning table to an FSC.}
	\label{fig:fsc-representation}
\end{wrapfigure}
A learning table is \emph{consistent} if for each $r_1,r_2\in \upperrows$ such that $r_1\equiv r_2$, we have $r_1\cdot e \equiv r_2\cdot e$ for all $e\in \observations$.
Closure of a learning table guarantees that each transition -- defined in the FSC by a lower row -- leads to a valid node, i.e.\ the node corresponding to the equivalent upper row.
Consistency, on the other hand, guarantees that the table unambiguously defines the output of a node in the FSC given an observation.

Using the notions of closure and concistency, we can define the transformation of a learning table into the learned FSC :
\begin{definition}[Learned FSC]
Given a closed and consistent learning table $\learningtable = (\upperrows, \columns, \learntable)$, we obtain a \emph{learned FSC} $\fsc_\learningtable = (\fscstates_\learningtable, \fscactionmap_\learningtable, \fsctransitionmap_\learningtable, \fsclearningtableinitstate)$ where:\\
$\fscstates_\learningtable= \{[r] \mid r\in\upperrows\}$, i.e., the nodes are the upper rows of the table;
$\fscactionmap_\learningtable([r],o) = \learntable(r,o)$ for all $o\in\observations$, i.e. the output of a transition is defined by its entry in the table;
$\fsctransitionmap_\learningtable([r],o) = [r \cdot o]$ for all $r \in \upperrows, o\in \observations$, i.e., the destination of a transition from node $[r]$ with observation $o$ is the node corresponding to the upper row equivalent to the lower row $r \cdot o$;
$\fsclearningtableinitstate = [\epsilon]$, i.e., the initial state is $[\epsilon]$.
\end{definition}

\begin{example}\label{ex:table-to-fsc}
	We demonstrate how to transform a table to an FSC in \cref{fig:fsc-representation}.
	The upper rows become states, the lower rows show the transitions.
	In this example, on both the observations {\color{OliveGreen}\texttt{x},\texttt{y}}, we stay in the  state and play action a and b, respectively.
\end{example}

\subsection{Algorithm}\label{sec:learning-fsc-alg}
We present our algorithm for learning an FSC from a strategy table. 
We have already seen the abstract view of the approach in \cref{fig:flow-chart}.
Algorithm~\ref{alg:learn-fsc} contains the pseudo-code for our learning algorithm. 
It consists of four main parts, also pictured in \cref{fig:flow-chart}: {\color{persianindigo}initialization}, {\color{lincolngreen}equivalence check}, {\color{mulberry}update of the FSC}, {\color{goldenpoppy}minimization}.

First, we initialise the learning table. The columns are initially filled with all available observations $\observations$, i.e. we set $\columns \leftarrow \observations$. 
We start with a single upper row $\epsilon$, representing the empty observation sequence. 
In the lower rows, we add the observation sequences of length $1$.
The entries of the table are then filled using output queries.
For example, consider the strategy table in \cref{tab:lookup-table}. The learning table after initialisation is shown in \cref{tab:example-init-table}.
The strategy table only contains observation sequences starting with \texttt{i}. 
Thus, for any sequence starting with \texttt{b} or \texttt{y}, all entries are $\dontcare$.

After initialising the table, we check whether it is closed.
If the table is not closed, all rows in the lower part of the table that do not occur in the upper part are moved to the upper part. Formally, we set $\upperrows \leftarrow \upperrows \cup \{l\}$ for all $l\in \upperrows\cdot \observations$ with $l \not\equiv u$ for all $u\in \upperrows$. 
In our example, this means that we move the rows $(\texttt{i}\;|\; \color{Blue}\dag  \; \color{Blue}\text{r}  \; \color{Blue}\text{d})$ and $(\texttt{b}\;|\; \color{Blue}\dontcare
\; \color{Blue}\dontcare \; \color{Blue}\dontcare)$ to the upper part of the table.
\scalebox{0.9}{
	\centering
	\begin{minipage}[t!]{\textwidth}
		\centering
		\vspace{-15pt}
		\begin{algorithm}[H]
			\centering
			\caption{Learning an FSC}\label{alg:learn-fsc}
			\begin{algorithmic}[1]
				\REQUIRE POMDP $\pomdp$, strategy table $\strategytable$
				\newline
				\algemph{algo-teal}{
					\STATE $\upperrows \gets \{\epsilon\}, \columns \gets \observations$
					\FORALL{$r\in\upperrows\cup\upperrows\cdot\observations, e\in\columns$}
					\STATE $\learntable(r, e)\gets$\Call{OutputQuery}{$r\cdot e$}
					\ENDFOR}
				\STATE \algemph{algo-teal}{\Call{MakeClosedAndConsistent}{$\upperrows, \columns, \learntable$}}
				\STATE \algemph{algo-lime}{$c\gets$\Call{EquivalenceQuery}{$\strategytable, \fsc_{(\upperrows, \columns, \learntable)}$}}
				\WHILE{$c\neq\epsilon$}
				\newline\algemph{algo-pink}{
					\STATE $\columns\gets \columns \ \cup $ set of all prefixes of $c$
					\FORALL{$r\in R\cup R\cdot\observations, e\in C$}
					\STATE $\learntable(r, e)\gets$\Call{OutputQuery}{$r\cdot e$}
					\ENDFOR}
				\STATE \algemph{algo-pink}{\Call{MakeClosedAndConsistent}{$\upperrows, \columns, \learntable$}}
				\STATE \algemph{algo-lime}{$c\gets$\Call{EquivalenceQuery}{$\strategytable, \fsc_{(\upperrows, \columns, \learntable)}$}}
				\ENDWHILE
				\STATE \algemph{algo-yellow}{$\learningtable \gets (\upperrows, \columns, \learntable)$,$\learningtable\gets$\Call{Minimize}{$\learningtable$}}
				\ENSURE FSC $\fsc_\learningtable$ generated from $\strategytable$
			\end{algorithmic}
		\end{algorithm}
	\end{minipage}
}\\

Once the table is closed (and naturally consistent),
we check for each row in the given strategy table $\strategytable$ whether it coincides with the action provided for this observation sequence by our hypothesis FSC $\fschypothesis$.
This is done formally using the equivalence query, i.e. we check if $EQ_\strategytable(\fschypothesis) = \epsilon$. 
If our hypothesis is not correct, we get a counterexample $c \in \observations^+$ where the output of $\strategytable$ and $\fschypothesis$ differ.
We add all non-empty prefixes of $c$ to $\columns$ and fill the table.
We repeat this until $\fschypothesis$ is equivalent to the strategy table $\strategytable$.

After the equivalence has been established, we use the ``don't-care" entries $\dontcare$ to further minimise the FSC.
These entries only appear for observation sequences that 
do not occur in the strategy table.
Thus, changing them to any action does not change the FSC's behaviour with respect to the strategy table.
We use this fact to merge nodes of the FSC to obtain a smaller one that still captures the behaviour of the strategy table.
It is not trivial to already exploit ``don't care" entries during the learning phase.
Two upper rows that are compatible in terms of the outputs they suggest, i.e. they either agree or have a $\dontcare$ where the other suggests an output, might be split when a new counterexample is added.
As such, we postpone minimisation of the FSC until the learning is finished.

\subsection{Proof of Concept: Belief Exploration}\label{sec:poc-belief-exploration}
For integrating our learning approach with an existing POMDP solution framework, we need to consider how the strategy table is constructed.
Assume that the solution method outputs \emph{some} representation of a strategy.
For strategies that are equivalent to some FSC, one possibility is to pre-compute the strategy table.
However, it is not clear how to determine the length of observation sequences that need to be considered.
A more reasonable view is considering the strategy representation as a \emph{symbolic} representation of the strategy table as long as it permits computable output and equivalence queries.

We demonstrate how this works by considering the belief exploration framework of \cite{bork2022under}.
The idea of belief exploration is to explore (a fragment of) the \emph{belief MDP} corresponding to the POMDP.
Then, model checking techniques are used on this finite MDP to optimise objectives and find a strategy.
States of the belief MDP are \emph{beliefs} -- distributions over states of the POMDP that describe the likelihood of being in a state given the observation history.
The strategy output of the belief exploration is a memoryless deterministic strategy $\mdpstrategy_{bel}$ that maps each belief to the optimal action.
It is well-known that there is a direct correspondence between strategies on the belief MDP and its POMDP \cite{smallwood1973}.
A decision in a belief corresponds to a decision in the POMDP for all observation sequences that lead to the belief in the belief MDP.
Thus, $\mdpstrategy_{bel}$ can also be interpreted as a strategy for the POMDP that we want to learn using our approach.

For now, assume that the belief MDP is finite.
Defining the computation of the output query is conceptually straightforward.
During each output query, we search for the belief $b$ that corresponds to the observation sequence in the belief MDP.
If we find it, the output is $\mdpstrategy_{bel}(b)$, otherwise the query outputs ``don't care" (\dontcare).
For the equivalence query, we consider one representative observation sequence for each belief $b$.
We compare whether $\mdpstrategy_{bel}(b)$ coincides with the output of the hypothesis FSC on the corresponding observation sequence.
If not, this sequence is a counterexample.

To deal with infinite belief MDPs, \cite{bork2022under} employs a partial exploration of the reachable belief space of the POMDP.
At the points where the exploration has been stopped (\emph{cut-off states}), they use approximations based on pre-computed, small strategies on the POMDP to yield a finite abstraction of the belief MDP.
The strategy $\mdpstrategy_{bel}$ computed on this abstraction, however, does not output valid actions for the POMDP in the cut-off states.
We modify the output query described above and introduce a set of $\dontknow$ symbols, i.e.,  $\dontknow_0,...\dontknow_n$.
On observation sequences of cut-off states, the output query returns ``don't-know" corresponding to that cutoff, i.e., $\dontknow_i$ for ``cut-off'' strategy $i$.
This allows us to later integrate the strategies used for approximation in our learned FSC or even substitute these strategies by different ones.

\subsection{Improving Learned FSCs for Incomplete Information}
\label{sec:cutoffHeuristics}
FSCs learned using the learning approach described in \Cref{sec:learning-fsc-alg} may still contain transitions with output ``don't-know" (\dontknow).
To make the FSC applicable to a POMDP, these outputs need to be replaced by distributions over actions of the POMDP.
For this purpose, we suggest two heuristics.
They are designed to be general, i.e. they do not consider any information that the underlying POMDP solution method provides.
Furthermore, they use the idea that already learned behavior might offer a basis for generalization.
As a result, the information already present in the FSC is used to replace the ``don't-know" outputs.
We note that additional heuristics can take for example the structure of the POMDP or information available in the POMDP solution method used to generate the strategy table into account.
For illustrating the heuristics, we assume that all output distributions are Dirac.
We denote the number of transitions in the FSC with observation $o$ with output not equal to $\dontcare_i$ or $\dontknow_i$ for some $i$ by $\#(o)$ and the number of transitions with output action $a$ for $o$ by $\#(o,a)$. 
\vspace{-10pt}
\begin{itemize}
\item \emph{Heuristic 1 -- Distribution:}
Intuitively, this heuristic replaces ``don't know'' by a distribution over all actions that the FSC already chooses for an observation. 
The resulting FSC therefore represents a randomized strategy, i.e. the strategy may probabilistically choose between actions.
This happens only in nodes of the FSC where ``don't know'' occurs.
Furthermore, this does not mean that the FSC itself is randomized; its structure remains deterministic.
Only some outputs represent randomization over actions.
In this method, we replace the $i$th ``don't know" $\dontknow_i$ by an action distribution where the probability of action $a$ under observation $o$ is given by $\frac{\#(o,a)}{\#(o)}$.
If $\#(o) = 0$, we keep $\dontknow_i$ instead which, in the belief exploration approach of \textsc{Storm}, represents a precomputed cutoff strategy.
In approaches, where the strategy does not provide any information at all, it can be replaced by $\dontcare$.
Intuitively, we try to copy the behavior of the FSC for this observation and since we don't know exactly which action would be optimal, we use a distribution over all possible actions.
\item \emph{Heuristic 2 -- Minimizing Using $\dontcare$-transitions:}
As for ease of implementation and explainability, smaller FSCs are preferable, this heuristic aims at replacing $\dontknow_i$ outputs such that we can minimise the FSC as much as possible.
For this purpose, we simply replace all occurrences of $\dontknow_i$ by $\dontcare$, i.e. we replace ``don't-know" by ``don't-care" outputs.
This allows the FSC to behave arbitrarily on these transitions. 
By then applying an additional minimisation step, we can potentially reduce the size of our FSC.
Intuitively, this allows for a smaller FSC that might be able to generalize better than specifying all actions directly.
Note that this heuristic will transform any deterministic FSC into a smaller representation that is still deterministic, and will not induce any randomization.
\end{itemize}

\section{Experimental Evaluation}\label{sec:experiments}
We implemented a prototype of the policy automaton learning framework on top of version 1.8.1 of the probabilistic model checker \tool{Storm} \cite{storm}. 
As input, our implementation takes the belief MC induced by the optimal policy on the belief MDP abstraction computed by \tool{Storm}'s belief exploration for POMDPs \cite{bork2022under}.
This Markov chain, labeled with observations and actions chosen by the computed strategy, encodes all information necessary for our approach as described in \Cref{sec:poc-belief-exploration}.
We apply our learning techniques to obtain a finite-state controller representation of a policy.
This FSC can be exported into a human-readable format or analyzed by building the Markov chain induced by the learned policy \emph{directly} on the POMDP.
As a baseline comparison for the learned FSC, we use the tool \tool{PAYNT} \cite{paynt}.
Recall that
\tool{PAYNT} uses a technique called \emph{inductive synthesis} to directly synthesize FSCs with respect to a given objective.

While recent research has shown that inductive synthesis and belief exploration improve when working in tandem \cite{andriushchenko2023}, we do not consider this symbiotic approach here.
However, we emphasize that integrating our approach in the framework of \cite{andriushchenko2023} is a promising prospect for future work.

\inlineheadingbf{Setup}
The experiments are run on two cores of an Intel\textsuperscript{\textregistered} Xeon\textsuperscript{\textregistered} Platinum 8160 CPU using 64GB RAM and a time limit of 1 hour.
We run \tool{Storm}'s POMDP model checking framework using default parameters.
In particular, we use the heuristically determined exploration depth for the belief MDP approximation and apply cut-offs where we choose not to explore further.
We refer to \cite{bork2022under} for more information.
For \tool{PAYNT}, we use abstraction-refinement with multi-core support \cite{andriushchenko2022}.
We run experiments for the two heuristics described in \Cref{sec:cutoffHeuristics}.
Additionally, we provide another result described as the ``base'' approach.
This is specific to the input given by \tool{Storm} and encodes the strategy obtained from \tool{Storm} exactly by keeping the cut-off strategies, represented as $\dontknow_i$ (see \cref{app:base-case} for more technical details).

\inlineheadingbf{Benchmarks}
As benchmarks for our evaluation, we consider the models from \cite{andriushchenko2023}. 
The benchmark set contains models taken from the literature \cite{andriushchenko2022,bork2020,bork2022under,hauskrecht1997} meant to illustrate the strengths and weaknesses of the belief exploration and inductive synthesis approaches.
As such, they also showcase how our learning approach transforms the output of the belief exploration concerning the size and quality of the computed FSC.
We provide an overview of the benchmarks, including their size, used in the paper in \cref{app:benchmarks}, \cref{tab:overview-benchmarks}.

\begin{figure}[t]
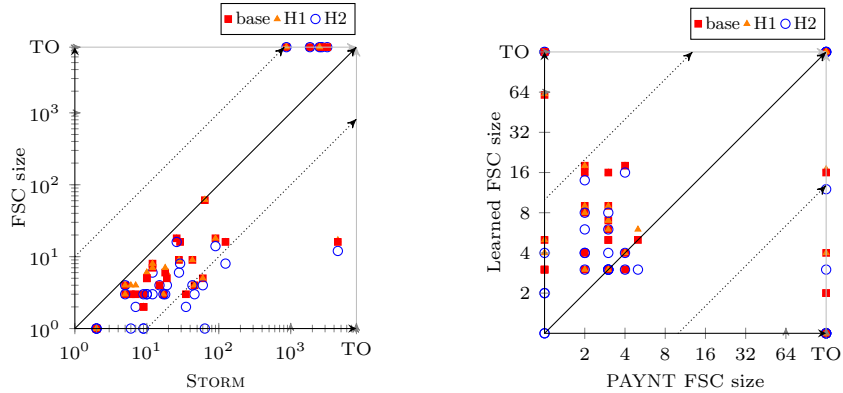

	\begin{subfigure}[t]{0.485\textwidth}
		\centering
		\setlength{\scatterplotsize}{0.9\textwidth}
		\scatterplotSizeStorm{data/storm-size.csv}{mc-size}{ \tool{Storm}}{automaton-size}{FSC size}{true}
		\caption{Size of input MC (from \tool{STORM}) vs.\ size of learned FSC (number of nodes)}
		\label{fig:size-comp-storm}
	\end{subfigure}
	\hfill
	\begin{subfigure}[t]{0.485\textwidth}
		\centering
		\setlength{\scatterplotsize}{0.9\textwidth}
		\scatterplotSize{data/second-table.csv}{paynt-size}{\tool{PAYNT} FSC size}{automaton-size}{Learned FSC size}{true}
		\caption{Size of learned FSC in comparison to \tool{PAYNT} (in number of nodes)}
		\label{fig:size-comp}
	\end{subfigure}
	
	\caption{Comparison of the resulting FSC size}
	\vspace{-17pt}
\end{figure}

\subsection{Results}\label{sec:results}
Our approach is general and meant to be used on top of other algorithms to transform possibly big and hardly explainable strategies into small FSCs.
However, we want to explore whether our results are comparable to state-of-the-art work for directly learning FSCs.
Therefore, we compare our FSCs to \tool{PAYNT}.

We present our results as follows: 
first, we talk about the size of the FSC generated by our method compared to the MC generated by \tool{Storm} and the FSCs generated by \tool{PAYNT}.
Secondly, we show the scalability of our approach by comparing the runtime with \tool{PAYNT}.
Lastly, we discuss the quality of the synthesized FSCs compared to \tool{PAYNT} and also discuss the trade-off between runtime and quality of the FSC.
\vspace{-8pt}

\subsubsection{Small and Explainable FSCs}
\label{sec:explainability}
\begin{wrapfigure}[10]{r}{0.35\textwidth}
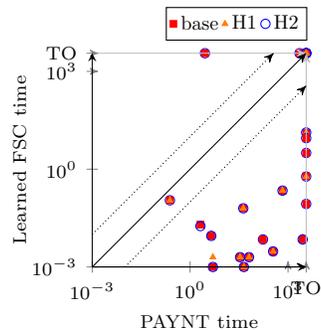

	\vspace{-70pt}
	\centering
	\scatterplotTime{data/second-2.csv}{paynt-time}{\tool{PAYNT} time}{automaton-time}{Learned FSC time}{true}
	\vspace{-15pt}
	\caption{Runtime comparison: our approach vs.\ \tool{PAYNT}}
	\label{fig:time-comp}
\end{wrapfigure}
Given a strategy table, our approach results in the \emph{smallest possible} FSC for the represented strategy.
As an overview, in \cref{fig:size-comp-storm}, we show a comparison of the sizes of the belief MC from \tool{Storm} to the size of our FSC.
The dashed line corresponds to a 10-fold reduction in size, showing our approach's usefulness.
We generate FSCs of sizes 1 to 64; however, more than 80\% of the FSCs are smaller than ten nodes, and only two are bigger than 60.
More than half of the generated FSCs have less than four nodes. 
In one case, we reduce 4517 states in the belief MC to an FSC of size 12.

We claim that these concise representations can generally be considered explainable, in particular when compared to huge original strategy representations.
When the given strategy is deterministic, 
our learning approach would construct a deterministic FSC which is easy to explain.
While improving the FSC by replacing the ``don't know'' actions (\Cref{sec:cutoffHeuristics}),
heuristic 2 still keeps the FSC deterministic
as it only replaces the $\dontknow$ actions with $\dontcare$ actions before minimisation.
Heuristic 1 often introduces some randomization when it replaces the $\dontknow$ actions with a distribution. 
But even in that case, they are
only in selected sink states which does not impede explainability.

In \cref{fig:size-comp}, we provide the size comparison of \tool{PAYNT}'s FSCs and ours.
Our FSCs are slightly bigger than \tool{PAYNT}'s in general, but our approach also returns smaller FSCs in some cases.
This is to be expected since the approach of \tool{PAYNT} is iteratively searching through the space of FSCs, starting with only one memory node and adding memory only once it is necessary.
Therefore, it is meant to find the smallest possible FSC.
However, \tool{PAYNT} times out much more often because of its exhaustive search on small FSC.
Additionally, our FSC are bound to be as big as necessary to represent the given strategy. 
Let us consider the benchmark \texttt{grid-avoid-4-0}. 
In this model, a robot moves in a grid of size four by four with no knowledge about its position. 
It starts randomly at any place in the grid and has to move towards a goal without falling into a ``hole''. 
\tool{PAYNT} produces a strategy of size 3, which moves right once and then iterates between moving right and down.
The nature of \tool{Storm}'s exploration leads to a strategy that moves right three times and then down forever. 
This can be represented in an FSC of size at least 5.

\vspace{-10pt}

\subsubsection{Scalability}
\label{sec:scalability}
Regarding scalability, \Cref{fig:time-comp} shows that our approach outperforms \textsc{PAYNT} on almost all cases. 
The dotted lines show difference by a factor of 10.
There are only two benchmarks, for which our approach times out and \textsc{PAYNT} does not. 
In one of these cases, \textsc{PAYNT} also takes more than 2000s to produce a result.

\vspace{-10pt}

\subsubsection{Runtime and Quality of FSCs}
\begin{table}[t]
	\caption{Comparison to PAYNT on value, size, and time (in that order) on selected benchmarks. The reported time for our approach includes the time of \tool{STORM} for producing the strategy table \textbf{and} the time for learning the FSC.}
	\centering
	\scalebox{0.8}{
		\renewcommand{\arraystretch}{0.95}
		\begin{tabular}{|c|c||c|c||c|c|c|}
			\hline
			&&&\multicolumn{3}{c|}{Learning heuristics}&\\
			Category & Model & \tool{Storm} & base & H$_1$ & H$_2$ & \tool{Paynt}\\
			\hline\hline
			
			\multirow{9}{*}{ \textbf{A}}
			&\multirow{2}{*}{problem-paynt-storm-combined}&8.07&8.07&\textbf{7.67}&\textbf{7.67}&\textbf{7.67}\\
			&&18&6&7&\textbf{3}&\textbf{3}\\
			&Rmin&\textless 1s&\multicolumn{3}{c|}{\textbf{\textless 1s}}&349s\\
			\cline{2-7}
			&\multirow{2}{*}{problem-storm-extended}&3009.0&3009.0&\textbf{98.0}&\textbf{98.0}&\textbf{98.0}\\
			&&64&61&62&\textbf{1}&\textbf{1}\\
			&Rmin&\textless 1s&\multicolumn{3}{c|}{\textbf{\textless 1s}}&\textbf{\textless 1 s}\\
			\cline{2-7}
			
			&\multirow{2}{*}{refuel-20}&0.14&0.14&\textbf{0.23}&\textbf{0.23}&TO\\
			&&46&4&4&\textbf{3}&\\
			&Pmax&73s&s
			75s&
			\textbf{74s}&
			\textbf{74s}&\\
			\cline{2-7}
			
			&\multirow{2}{*}{grid-avoid-4-01}&0.75&0.75&0.9&0.67&\textbf{0.93}\\
			&&10&5&6&\textbf{3}&5\\
			&Pmax&\textless 1s&\multicolumn{3}{c|}{\textbf{\textless 1s}}&726s\\
			\hline

			\multirow{6}{*}{ \textbf{B}}
			&\multirow{2}{*}{posterior-awareness}&12.0&12.0&12.0&12.0&\textbf{11.99}\\
			&&5&\textbf{4}&\textbf{4}&\textbf{4}&\textbf{4}\\
			&Rmin&\textless 1s&\multicolumn{3}{c|}{\textbf{\textless 1s}}&\textbf{\textless 1 s}\\
			\cline{2-7}
			&\multirow{2}{*}{4x5x2-95}&1.29&1.29&1.28&1.26&\textbf{2.02}\\
			&&26&18&18&16&\textbf{4}\\
			&Rmax&\textless 1s&\multicolumn{3}{c|}{\textbf{\textless 1s}}&2807s\\
			\hline

			\multirow{6}{*}{ \textbf{C}}
			&\multirow{2}{*}{query-s2}&395.66&395.66&391.9&343.94&\textbf{486.69}\\
			&&43&9&9&4&\textbf{2}\\
			&Rmax&\textless 1s&\multicolumn{3}{c|}{\textbf{\textless 1s}}&5s\\
			\cline{2-7}
			
			&\multirow{2}{*}{drone-4-1}&0.75&TO&TO&TO&\textbf{0.87}\\
			&&3217&&&&\textbf{1}\\
			&Pmax&1s&&&&\textbf{2250s}\\
			\hline
			
		\end{tabular}
	}
	\vspace{-3pt}
\end{table}

Comparing the quality of results, we need to put into consideration that our approach often runs within a fraction of the available time.
We run \tool{Storm} with its default values to get a strategy.
As demonstrated in \cite{andriushchenko2022}, running \tool{Storm} using non-default parameters, specifically larger exploration thresholds, results in better strategies at the cost of longer runtimes.
Our approach directly profits from such better input strategies.

Since the learning is done in far less than a second for most of the benchmarks, we suggest using a portfolio of the heuristics.
This allows us to output the optimal solution among all our heuristics with negligible computational overhead.
To simplify the presentation of our results, we categorize the benchmarks into three groups: A, B, and C, based on the overall performance of our method. 
Due to space constraints, we provide detailed results for only a selection of benchmarks for each category and do not discuss benchmarks for which both approaches experienced timeouts. 
The complete set of results is given in \cref{app:results-table}. \\

\noindent
\emph{Category A.}
This category represents benchmarks where our approach is arguably favored, assuming the portfolio approach. 
There are a total of 19 benchmarks in this category, and we observe that we can improve all variants of properties using heuristics. 
Only one time, \tool{PAYNT} produces a slightly better probability value ($0.93$ vs $0.9$), but it takes significantly more time ($726$s vs $<1$s).
There are 7 cases where we can generate FSCs while \tool{PAYNT} times out and on 6 out of these 7 cases, we get the \textbf{smallest} FSCs reported in state-of-the-art \cite{andriushchenko2023}.
In this category, we also include benchmarks on which the heuristics improved on \textsc{Storm}'s strategy to achieve the same value as \tool{PAYNT} while being more efficient, e.g. \texttt{problem-paynt-storm-combined}.
Also, for the benchmark \texttt{problem-storm-extended}, designed to be difficult for \tool{Storm}, we reduce the approximate total reward from $3009$ to $98$, resulting in an FSC of size \textbf{1} in \textbf{$<1$s}.

\vspace{5pt}
\noindent
\emph{Category B.}
This category corresponds to the benchmarks on which there is no clear front-runner.
There are a total of 7 benchmarks in this category.
Three of these benchmarks are similar to \texttt{posterior-awareness}, where the results produced and the time taken are quite similar for both approaches.
The other 4 benchmarks (similar to \texttt{4x5x2-95}) show that the value generated by our approach is significantly worse; however, it takes significantly less time.
Depending on the situation, this trade-off between quality and runtime may favor either approach.

\vspace{5pt}
\noindent
\emph{Category C.}
This category shows the weakness of our method compared to \tool{PAYNT}. 
In this category, there are a total of 3
benchmarks, out of which our approach times out 2 times. 
It is notable that the \texttt{drone}-benchmarks seem to be generally hard: \tool{PAYNT} needs 2250s for \texttt{drone-4-1}, and both approaches time out for the bigger instances.
There is only one benchmark, \texttt{query-s2}, where we produce a worse value without any significant time advantage over \tool{PAYNT}.

\vspace{-10pt}
\section{Conclusion}
In this paper, we present an approach to learn an FSC for representing POMDP strategies.
Our FSCs are (i) always \emph{smaller} than the given representation, and (ii) the FSC structure is simple, which together makes the strategy more \emph{explainable}.
Indeed, concerning the structure has the required desirable properties.
First, the structure of the FSC is always deterministic.
Second, one of our heuristics only generates deterministic output actions (without randomization).
While the other heuristic can lead to the representation of a randomized strategy, it only randomizes the output actions, \emph{not} the FSC structure.
Besides, the randomization happens in only a very restricted form
as discussed in \Cref{sec:explainability}.

Further, our heuristics achieved notable improvements in the \emph{performance} of many strategies produced by \tool{Storm} and provably perform equal or better than the baseline, while retaining negligible resource consumption.
Altogether, our comparative analysis against \tool{PAYNT} underscores the competitiveness of our proposed method, frequently yielding FSCs of comparable quality with incomparable scalability. 
Specifically, for a set of six benchmarks where \tool{PAYNT} reaches its computational limits, we have constructed 
the smallest FSCs reported in the existing literature. 
This not only attests to the scalability and efficiency of our approach but also highlights its applicability in scenarios challenging for other tools.

Concerning future work, several directions open up.
Obviously, further heuristics can be designed to solve some of the patterns occurring in the cases where our approach could not match the size achieved by PAYNT.
Further, besides applying our approach to strategies generated by other approaches, we would like to integrate it into the other approaches in order to improve them.

\paragraph*{Data Availability.} The artifact accompanying this paper \cite{artifact} contains source code, benchmark files, and replication scripts for our experiments.

\bibliographystyle{splncs04}
\bibliography{ref}

\begin{thebibliography}{10}
\providecommand{\url}[1]{\texttt{#1}}
\providecommand{\urlprefix}{URL }
\providecommand{\doi}[1]{https://doi.org/#1}

\bibitem{amato2010}
Amato, C., Bernstein, D.S., Zilberstein, S.: Optimizing fixed-size stochastic
  controllers for pomdps and decentralized pomdps. Auton. Agents Multi Agent
  Syst.  \textbf{21}(3),  293--320 (2010),
  \url{https://doi.org/10.1007/s10458-009-9103-z}

\bibitem{andriushchenko2023}
Andriushchenko, R., Bork, A., Ceska, M., Junges, S., Katoen, J., Mac{\'{a}}k,
  F.: Search and explore: Symbiotic policy synthesis in pomdps. In: Computer
  Aided Verification - 35th International Conference, {CAV} 2023, Paris,
  France, July 17-22, 2023, Proceedings, Part {III}. Lecture Notes in Computer
  Science, vol. 13966, pp. 113--135. Springer (2023),
  \url{https://doi.org/10.1007/978-3-031-37709-9\_6}

\bibitem{andriushchenko2022}
Andriushchenko, R., Ceska, M., Junges, S., Katoen, J.: Inductive synthesis of
  finite-state controllers for pomdps. In: Uncertainty in Artificial
  Intelligence, Proceedings of the Thirty-Eighth Conference on Uncertainty in
  Artificial Intelligence, {UAI} 2022, 1-5 August 2022, Eindhoven, The
  Netherlands. Proceedings of Machine Learning Research, vol.~180, pp. 85--95.
  {PMLR} (2022),
  \url{https://proceedings.mlr.press/v180/andriushchenko22a.html}

\bibitem{paynt}
Andriushchenko, R., Ceska, M., Junges, S., Katoen, J., Stupinsk{\'{y}}, S.:
  {PAYNT:} {A} tool for inductive synthesis of probabilistic programs. In:
  Computer Aided Verification - 33rd International Conference, {CAV} 2021,
  Virtual Event, July 20-23, 2021, Proceedings, Part {I}. Lecture Notes in
  Computer Science, vol. 12759, pp. 856--869. Springer (2021).
  \doi{10.1007/978-3-030-81685-8\_40},
  \url{https://doi.org/10.1007/978-3-030-81685-8\_40}

\bibitem{angluin1987learning}
Angluin, D.: Learning regular sets from queries and counterexamples.
  Information and computation  \textbf{75}(2),  87--106 (1987),
  \url{https://doi.org/10.1016/0890-5401(87)90052-6}

\bibitem{DBLP:conf/hybrid/AshokJJKWZ20}
Ashok, P., Jackermeier, M., Jagtap, P., Kret{\'{\i}}nsk{\'{y}}, J., Weininger,
  M., Zamani, M.: dtcontrol: decision tree learning algorithms for controller
  representation. In: {HSCC}. pp. 17:1--17:7. {ACM} (2020),
  \url{https://dl.acm.org/doi/abs/10.1145/3365365.3383468}

\bibitem{DBLP:conf/tacas/AshokJKWWY21}
Ashok, P., Jackermeier, M., K{\v r}et{\'{\i}}nsk{\'{y}}, J., Weinhuber, C.,
  Weininger, M., Yadav, M.: dtcontrol 2.0: Explainable strategy representation
  via decision tree learning steered by experts. In: {TACAS} {(2)}. Lecture
  Notes in Computer Science, vol. 12652, pp. 326--345. Springer (2021),
  \url{https://doi.org/10.1007/978-3-030-72013-1_17}

\bibitem{artifact}
Bork, A., Chakraborty, D., Grover, K., Mohr, S., Kretinsky, J.: {Artifact for
  Paper: Learning Explainable and Better Performing Representations of POMDP
  Strategies}, \url{https://doi.org/10.5281/zenodo.10437018}

\bibitem{bork2020}
Bork, A., Junges, S., Katoen, J., Quatmann, T.: Verification of
  indefinite-horizon pomdps. In: Automated Technology for Verification and
  Analysis - 18th International Symposium, {ATVA} 2020, Hanoi, Vietnam, October
  19-23, 2020, Proceedings. Lecture Notes in Computer Science, vol. 12302, pp.
  288--304. Springer (2020),
  \url{https://doi.org/10.1007/978-3-030-59152-6\_16}

\bibitem{bork2022under}
Bork, A., Katoen, J.P., Quatmann, T.: Under-approximating expected total
  rewards in pomdps. In: International Conference on Tools and Algorithms for
  the Construction and Analysis of Systems. pp. 22--40. Springer (2022),
  \url{https://doi.org/10.1007/978-3-030-99527-0_2}

\bibitem{DBLP:conf/cav/BrazdilCCFK15}
Br{\'{a}}zdil, T., Chatterjee, K., Chmelik, M., Fellner, A., K{\v
  r}et{\'{\i}}nsk{\'{y}}, J.: Counterexample explanation by learning small
  strategies in markov decision processes. In: {CAV} {(1)}. Lecture Notes in
  Computer Science, vol.~9206, pp. 158--177. Springer (2015),
  \url{https://doi.org/10.1007/978-3-319-21690-4_10}

\bibitem{carr2019}
Carr, S., Jansen, N., Wimmer, R., Serban, A.C., Becker, B., Topcu, U.:
  Counterexample-guided strategy improvement for pomdps using recurrent neural
  networks. In: Kraus, S. (ed.) Proceedings of the Twenty-Eighth International
  Joint Conference on Artificial Intelligence, {IJCAI} 2019, Macao, China,
  August 10-16, 2019. pp. 5532--5539. ijcai.org (2019),
  \url{https://doi.org/10.24963/ijcai.2019/768}

\bibitem{chatterjee2016decidable}
Chatterjee, K., Chmelik, M., Tracol, M.: What is decidable about partially
  observable markov decision processes with $\omega$-regular objectives.
  Journal of Computer and System Sciences  \textbf{82}(5),  878--911 (2016),
  \url{https://doi.org/10.1016/j.jcss.2016.02.009}

\bibitem{cubuktepe2021}
Cubuktepe, M., Jansen, N., Junges, S., Marandi, A., Suilen, M., Topcu, U.:
  Robust finite-state controllers for uncertain pomdps. In: Thirty-Fifth {AAAI}
  Conference on Artificial Intelligence, {AAAI} 2021, Thirty-Third Conference
  on Innovative Applications of Artificial Intelligence, {IAAI} 2021, The
  Eleventh Symposium on Educational Advances in Artificial Intelligence, {EAAI}
  2021, Virtual Event, February 2-9, 2021. pp. 11792--11800. {AAAI} Press
  (2021), \url{https://doi.org/10.1609/aaai.v35i13.17401}

\bibitem{hansen1998}
Hansen, E.A.: Solving pomdps by searching in policy space. In: Cooper, G.F.,
  Moral, S. (eds.) {UAI} '98: Proceedings of the Fourteenth Conference on
  Uncertainty in Artificial Intelligence, University of Wisconsin Business
  School, Madison, Wisconsin, USA, July 24-26, 1998. pp. 211--219. Morgan
  Kaufmann (1998), \url{https://dl.acm.org/doi/abs/10.5555/2074094.2074119}

\bibitem{hauskrecht1997}
Hauskrecht, M.: Incremental methods for computing bounds in partially
  observable markov decision processes. In: Proceedings of the Fourteenth
  National Conference on Artificial Intelligence and Ninth Innovative
  Applications of Artificial Intelligence Conference, {AAAI} 97, {IAAI} 97,
  July 27-31, 1997, Providence, Rhode Island, {USA}. pp. 734--739. {AAAI} Press
  / The {MIT} Press (1997),
  \url{https://dl.acm.org/doi/10.5555/1867406.1867520}

\bibitem{hauskrecht2000}
Hauskrecht, M.: Value-function approximations for partially observable markov
  decision processes. J. Artif. Intell. Res.  \textbf{13},  33--94 (2000),
  \url{https://doi.org/10.1613/jair.678}

\bibitem{heck2022}
Heck, L., Spel, J., Junges, S., Moerman, J., Katoen, J.: Gradient-descent for
  randomized controllers under partial observability. In: Verification, Model
  Checking, and Abstract Interpretation - 23rd International Conference,
  {VMCAI} 2022, Philadelphia, PA, USA, January 16-18, 2022, Proceedings.
  Lecture Notes in Computer Science, vol. 13182, pp. 127--150. Springer (2022),
  \url{https://doi.org/10.1007/978-3-030-94583-1\_7}

\bibitem{storm}
Hensel, C., Junges, S., Katoen, J., Quatmann, T., Volk, M.: The probabilistic
  model checker storm. Int. J. Softw. Tools Technol. Transf.  \textbf{24}(4),
  589--610 (2022), \url{https://doi.org/10.1007/s10009-021-00633-z}

\bibitem{junges2018}
Junges, S., Jansen, N., Wimmer, R., Quatmann, T., Winterer, L., Katoen, J.,
  Becker, B.: Finite-state controllers of pomdps using parameter synthesis. In:
  Globerson, A., Silva, R. (eds.) Proceedings of the Thirty-Fourth Conference
  on Uncertainty in Artificial Intelligence, {UAI} 2018, Monterey, California,
  USA, August 6-10, 2018. pp. 519--529. {AUAI} Press (2018)

\bibitem{KAELBLING199899}
Kaelbling, L.P., Littman, M.L., Cassandra, A.R.: Planning and acting in
  partially observable stochastic domains. Artificial Intelligence
  \textbf{101}(1),  99--134 (1998),
  \url{https://doi.org/10.1016/S0004-3702(98)00023-X}

\bibitem{kurniawati2008}
Kurniawati, H., Hsu, D., Lee, W.S.: {SARSOP:} efficient point-based {POMDP}
  planning by approximating optimally reachable belief spaces. In: Brock, O.,
  Trinkle, J., Ramos, F. (eds.) Robotics: Science and Systems IV,
  Eidgen{\"{o}}ssische Technische Hochschule Z{\"{u}}rich, Zurich, Switzerland,
  June 25-28, 2008. The {MIT} Press (2008),
  \url{https://doi.org/10.15607/RSS.2008.IV.009}

\bibitem{prism}
Kwiatkowska, M.Z., Norman, G., Parker, D.: {PRISM} 4.0: Verification of
  probabilistic real-time systems. In: Gopalakrishnan, G., Qadeer, S. (eds.)
  Computer Aided Verification - 23rd International Conference, {CAV} 2011,
  Snowbird, UT, USA, July 14-20, 2011. Proceedings. Lecture Notes in Computer
  Science, vol.~6806, pp. 585--591. Springer (2011),
  \url{https://doi.org/10.1007/978-3-642-22110-1\_47}

\bibitem{madani2003undecidability}
Madani, O., Hanks, S., Condon, A.: On the undecidability of probabilistic
  planning and related stochastic optimization problems. Artificial
  Intelligence  \textbf{147}(1-2),  5--34 (2003),
  \url{https://doi.org/10.1016/S0004-3702(02)00378-8}

\bibitem{meuleau1999}
Meuleau, N., Kim, K., Kaelbling, L.P., Cassandra, A.R.: Solving pomdps by
  searching the space of finite policies. In: Laskey, K.B., Prade, H. (eds.)
  {UAI} '99: Proceedings of the Fifteenth Conference on Uncertainty in
  Artificial Intelligence, Stockholm, Sweden, July 30 - August 1, 1999. pp.
  417--426. Morgan Kaufmann (1999),
  \url{https://dl.acm.org/doi/10.5555/2073796.2073844}

\bibitem{DBLP:conf/tacas/NeiderT16}
Neider, D., Topcu, U.: An automaton learning approach to solving safety games
  over infinite graphs. In: {TACAS}. Lecture Notes in Computer Science,
  vol.~9636, pp. 204--221. Springer (2016),
  \url{https://doi.org/10.1007/978-3-662-49674-9_12}

\bibitem{norman2017}
Norman, G., Parker, D., Zou, X.: Verification and control of partially
  observable probabilistic systems. Real Time Syst.  \textbf{53}(3),  354--402
  (2017), \url{https://doi.org/10.1007/s11241-017-9269-4}

\bibitem{pineau2003}
Pineau, J., Gordon, G.J., Thrun, S.: Point-based value iteration: An anytime
  algorithm for pomdps. In: Gottlob, G., Walsh, T. (eds.) IJCAI-03, Proceedings
  of the Eighteenth International Joint Conference on Artificial Intelligence,
  Acapulco, Mexico, August 9-15, 2003. pp. 1025--1032. Morgan Kaufmann (2003)

\bibitem{russell2010artificial}
Russell, S.J.: Artificial intelligence a modern approach. Pearson Education,
  Inc. (2010), \url{https://dl.acm.org/doi/book/10.5555/1671238}

\bibitem{mealymachines}
Shahbaz, M., Groz, R.: Inferring mealy machines. In: Cavalcanti, A., Dams, D.
  (eds.) {FM} 2009: Formal Methods, Second World Congress, Eindhoven, The
  Netherlands, November 2-6, 2009. Proceedings. Lecture Notes in Computer
  Science, vol.~5850, pp. 207--222. Springer (2009),
  \url{https://doi.org/10.1007/978-3-642-05089-3\_14}

\bibitem{shani2013survey}
Shani, G., Pineau, J., Kaplow, R.: A survey of point-based pomdp solvers.
  Autonomous Agents and Multi-Agent Systems  \textbf{27},  1--51 (2013),
  \url{https://doi.org/10.1007/s10458-012-9200-2}

\bibitem{aaai23safe}
Sim\~{a}o, T.D., Suilen, M., Jansen, N.: Safe policy improvement for pomdps via
  finite-state controllers. In: Proceedings of the Thirty-Seventh AAAI
  Conference on Artificial Intelligence and Thirty-Fifth Conference on
  Innovative Applications of Artificial Intelligence and Thirteenth Symposium
  on Educational Advances in Artificial Intelligence. AAAI'23/IAAI'23/EAAI'23,
  AAAI Press (2023), \url{https://doi.org/10.1609/aaai.v37i12.26763}

\bibitem{smallwood1973}
Smallwood, R.D., Sondik, E.J.: The optimal control of partially observable
  markov processes over a finite horizon. Oper. Res.  \textbf{21}(5),
  1071--1088 (1973), \url{https://doi.org/10.1287/opre.21.5.1071}

\bibitem{spaan2005}
Spaan, M.T.J., Vlassis, N.: Perseus: Randomized point-based value iteration for
  pomdps. J. Artif. Intell. Res.  \textbf{24},  195--220 (2005),
  \url{https://doi.org/10.1613/jair.1659}

\bibitem{thomas2015}
Thomas, P., Theocharous, G., Ghavamzadeh, M.: High-confidence off-policy
  evaluation. In: Proceedings of the AAAI Conference on Artificial
  Intelligence. vol.~29 (2015),
  \url{https://dl.acm.org/doi/10.5555/2888116.2888134}

\end{thebibliography}

\appendix
\section*{Appendix}
\section{Appendix}
\subsection{The ``base'' case - Exact Representation of \tool{Storm}'s strategy}\label{app:base-case}
Recall that the output query returns ``don't-know'' ($\dontknow$) for cut-off states.
In the ``base'' case, we introduce several actions $\dontknow_0,...\dontknow_n$, each indicating the specific cut-off strategy chosen in that state.
In other words, the strategy table contains a distinct symbol for each cut-off strategy.
Thus, our learning table will distinguish between them.
We treat these actions by only following cut-off strategy $i$ after the FSC has output $\dontknow_i$ for the first time.

\newpage
\subsection{Overview of the Benchmarks}\label{app:benchmarks}
\begin{table}
	\centering
	\caption{Overview of the used benchmarks. For each model, we show the number of states $|\mdpstates|$ of the POMDP, the number of transitions $\sum\mdpactions$, and the number of observations $|\observations|$.}
	\label{tab:overview-benchmarks}
	\scalebox{0.9}{\begin{tabular}{|c|ccc||c|ccc|}
\hline
Model & $|\mdpstates|$ & $\sum \mdpactions$ & $|\observations|$ &
	Model & $|\mdpstates|$ & $\sum \mdpactions$ & $|\observations|$
\\\hline
4x3-95&22&82&9 &
	network-3-8-20&17253&30597&2205\\
4x5x2-95&79&310&7  &
	network-prio-2-8-20&19373&34157&4909\\
drone-4-1&1226&2954&384 &
	posterior-awareness&5&9&3\\
drone-4-2&1226&2954&761 &
	problem-paynt&9&33&5\\
drone-8-2&13042&32242&3195 &
	problem-paynt-storm-combined&28&122&13\\
grid-avoid-4-0&17&59&4 &
	problem-storm&4&6&3\\
grid-avoid-4-01&17&59&4 &
	problem-storm-extended&101&199&3\\
grid-large-10-5&100&397&5 &
	problem-storm-paynt-combined&21&109&7\\
grid-large-20-5&400&1597&17 &
	query-s2&36&70&6\\
grid-large-30-5&900&3597&37 &
	query-s3&108&320&6\\
hallway2&1500&7492&20 &
	refuel-06&208&565&50\\
lanes-100-combined-new&2741&5285&11 &
	refuel-08&470&1431&66\\
maze-alex&15&54&8 &
	refuel-20&6834&24763&174\\
milos-aaai97&165&980&1 &
	rocks-12&6553&31537&1645\\
mini-hall2&27&77&12 &
	rocks-16&11017&54289&2761\\
network&19&70&5 &
	stand-tiger-95&14&50&7\\
network-2-8-20&4589&6973&1173 &
	web-mall&8&20&5\\
\hline
\end{tabular}}

\end{table}
\FloatBarrier
\newpage
\subsection{Full Results}\label{app:results-table}

\begin{figure}[h]
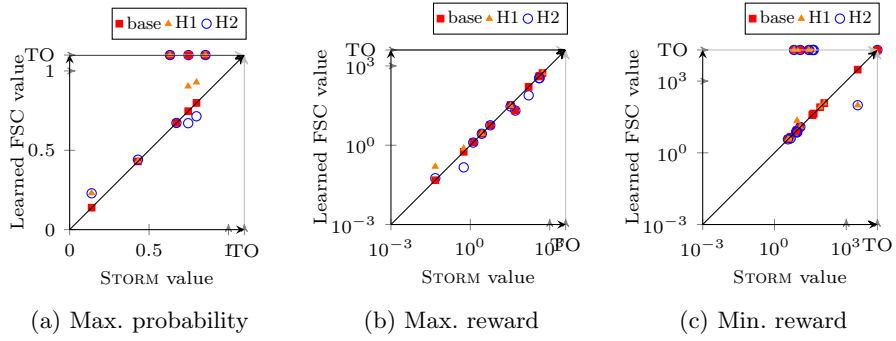

	\begin{subfigure}[t]{0.32\textwidth}
		\centering
		\setlength{\scatterplotsize}{\textwidth}
		\scatterplotProb{data/pmax.csv}{storm-result}{\textsc{Storm} value}{automaton-result}{Learned FSC value}{true}
		\caption{Max. probability}
		\label{pmax-comp-storm}
	\end{subfigure}
	\hfill
	\begin{subfigure}[t]{0.32\textwidth}
		\setlength{\scatterplotsize}{\textwidth}
		\centering
		\scatterplotRmax{data/rmax.csv}{storm-result}{\textsc{Storm} value}{automaton-result}{Learned FSC value}{true}
		\caption{Max. reward}
		\label{fig:rmax-comp-storm}
	\end{subfigure}
	\hfill
	\begin{subfigure}[t]{0.32\textwidth}
		\centering
		\setlength{\scatterplotsize}{\textwidth}
		\scatterplotRmin{data/rmin.csv}{storm-result}{\textsc{Storm} value}{automaton-result}{Learned FSC value}{true}
		\caption{Min. reward}
		\label{fig:rmin-comp-storm}
	\end{subfigure}
	\caption{Scatter plots of values generated by learned FSCs compared to the original result form \textsc{Storm} on different types of properties.}
	\label{fig:value-comp-storm}
\end{figure}

\begin{figure}[h]
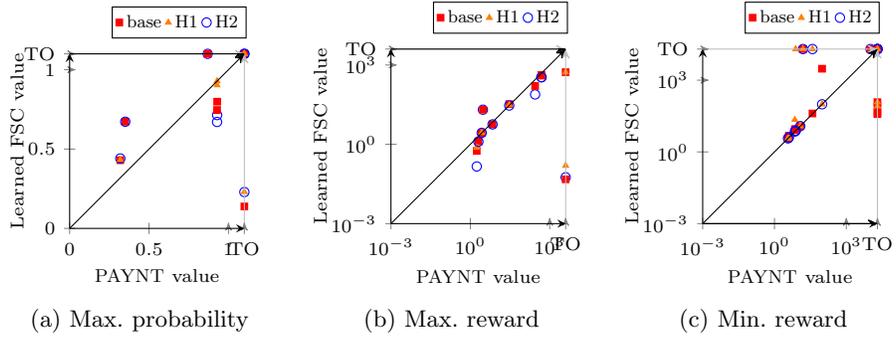

	\begin{subfigure}[t]{0.32\textwidth}
		\centering
		\setlength{\scatterplotsize}{\textwidth}
		\scatterplotProb{data/pmax.csv}{paynt-value}{\textsc{PAYNT} value}{automaton-result}{Learned FSC value}{true}
		\caption{Max. probability}
		\label{pmax-comp}
	\end{subfigure}
	\hfill
	\begin{subfigure}[t]{0.32\textwidth}
		\setlength{\scatterplotsize}{\textwidth}
		\centering
		\scatterplotRmax{data/rmax.csv}{paynt-value}{\textsc{PAYNT} value}{automaton-result}{Learned FSC value}{true}
		\caption{Max. reward}
		\label{fig:rmax-comp}
	\end{subfigure}
	\hfill
	\begin{subfigure}[t]{0.32\textwidth}
		\centering
		\setlength{\scatterplotsize}{\textwidth}
		\scatterplotRmin{data/rmin.csv}{paynt-value}{\textsc{PAYNT} value}{automaton-result}{Learned FSC value}{true}
		\caption{Min. reward}
		\label{fig:rmin-comp}
	\end{subfigure}
	\caption{Scatter plots of values generated by learned FSCs in comparison to \textsc{PAYNT} on different types of properties.}
	\label{fig:comp-value-paynt}
\end{figure}

\begin{small}
	\begin{longtable}{|c||c||c|c|c|c|}
	\caption{\textbf{Results on Category A Benchmarks}.
		Comparison to PAYNT on value, size, and time (in that order) on selected benchmarks. The reported time for our approach includes the time of \tool{STORM} for producing the strategy table \textbf{and} the time for learning the FSC.
		We produce the best value or the heuristics made improvements to achieve similar value as \tool{PAYNT}. 
		Among these, \tool{PAYNT} is better but only slightly on grid-avoid-4-01, for which we scale much better. 
		The table is partitioned into three parts according to the property (Rmin, Pmax, Rmax).
	}\label{tab:full-category-A}
	\endfirsthead
	\endhead
	\hline
	&&\multicolumn{3}{c|}{Learning heuristics}&\\
	Model & \textsc{Storm} & base & H$_1$ & H$_2$ & \textsc{Paynt}\\
	\hline\hline

	\multirow{2}{*}{problem-paynt-storm-combined}&8.07&8.07&\textbf{7.67}&\textbf{7.67}&\textbf{7.67}\\
	&18&6&7&\textbf{3}&\textbf{3}\\
	Rmin&\textless 1s&\multicolumn{3}{c|}{\textbf{\textless 1s}}&349s\\
	\hline
	\multirow{2}{*}{problem-storm-paynt-combined}&8.98&8.98&inf&\textbf{7.67}&\textbf{7.67}\\
	&12&7&7&\textbf{3}&\textbf{3}\\
	Rmin&\textless 1s&\multicolumn{3}{c|}{\textbf{\textless 1s}}&44s\\
	\hline
	\multirow{2}{*}{maze-alex}&8.77&8.77&22.2&\textbf{7.09}&\textbf{7.09}\\
	&29&16&16&8&\textbf{3}\\
	Rmin&\textless 1s&\multicolumn{3}{c|}{\textbf{\textless 1s}}&4s\\
	\hline
	\multirow{2}{*}{problem-storm-extended}&3009.0&3009.0&\textbf{98.0}&\textbf{98.0}&\textbf{98.0}\\
	&64&61&62&\textbf{1}&\textbf{1}\\
	Rmin&\textless 1s&\multicolumn{3}{c|}{\textbf{\textless 1s}}&\textbf{\textless 1 s}\\
	\hline
	\multirow{2}{*}{problem-storm}&4.75&4.75&\textbf{4.0}&\textbf{4.0}&\textbf{4.0}\\
	&6&3&4&\textbf{1}&\textbf{1}\\
	Rmin&\textless 1s&\multicolumn{3}{c|}{\textbf{\textless 1s}}&\textbf{\textless 1 s}\\
	\hline
	\multirow{2}{*}{grid-large-20-5}&80.51&\textbf{80.51}&\textbf{80.51}&ERROR&TO\\
	&2&\textbf{1}&\textbf{1}&\textbf{1}&\\
	Rmin&\textless 1s&\multicolumn{3}{c|}{\textbf{\textless 1s}}&\\
	\hline
	\multirow{2}{*}{grid-large-30-5}&120.52&\textbf{120.52}&\textbf{120.52}&ERROR&TO\\
	&2&\textbf{1}&\textbf{1}&\textbf{1}&\\
	Rmin&1s&\multicolumn{3}{c|}{
		\textbf{
		1s
	}}&\\
	\hline
	\multirow{2}{*}{rocks-12}&38.0&\textbf{38.0}&\textbf{38.0}&inf&TO\\
	&9&2&2&\textbf{1}&\\
	Rmin&1s&
	13s&
	\textbf{
		10s
	}&\textbf{
	10s}&\\
	\hline
	\multirow{2}{*}{rocks-16}&44.0&\textbf{44.0}&\textbf{44.0}&inf&TO\\
	&9&2&2&\textbf{1}&\\
	Rmin&2s&\textbf{
	34s}&
	38s&
	36s&\\
	\hline
	\hline

	\multirow{2}{*}{refuel-20}&0.14&0.14&\textbf{0.23}&\textbf{0.23}&TO\\
	&46&4&4&\textbf{3}&\\
	Pmax&73s&
	75s&\textbf{
	74s}&\textbf{
	74s}&\\
	\hline
	\multirow{2}{*}{refuel-08}&0.43&0.43&0.43&\textbf{0.44}&0.32\\
	&60&5&5&4&\textbf{1}\\
	Pmax&\textless 1s&\multicolumn{3}{c|}{\textbf{\textless 1s}}&679s\\
	\hline
	\multirow{2}{*}{grid-avoid-4-0}&0.8&0.8&\textbf{0.93}&0.71&\textbf{0.93}\\
	&10&5&6&\textbf{3}&\textbf{3}\\
	Pmax&\textless 1s&\multicolumn{3}{c|}{\textbf{\textless 1s}}&\textbf{\textless 1 s}\\
	\hline
	
	\multirow{2}{*}{refuel-06}&0.67&\textbf{0.67}&\textbf{0.67}&\textbf{0.67}&0.35\\
	&35&3&3&2&\textbf{1}\\
	Pmax&\textless 1s&\multicolumn{3}{c|}{\textbf{\textless 1s}}&2s\\
	\hline
	
	\multirow{2}{*}{grid-avoid-4-01}&0.75&0.75&0.9&0.67&\textbf{0.93}\\
	&10&5&6&\textbf{3}&5\\
	Pmax&\textless 1s&\multicolumn{3}{c|}{\textbf{\textless 1s}}&726s\\
	\hline
	\hline

	\multirow{2}{*}{hallway2}&0.05&0.05&\textbf{0.16}&0.06&TO\\
	&4517&16&17&\textbf{12}&\\
	Rmax&9s&
	15s&
	15s&\textbf{
	13s}&\\
	\hline

	\multirow{2}{*}{milos-aaai97}&33.14&\textbf{33.14}&30.94&29.31&29.46\\
	&7&3&4&2&\textbf{1}\\
	Rmax&\textless 1s&\multicolumn{3}{c|}{\textbf{\textless 1s}}&5s\\
	\hline

	\multirow{2}{*}{mini-hall2}&2.71&\textbf{2.71}&\textbf{2.71}&\textbf{2.71}&2.69\\
	&17&3&3&3&\textbf{2}\\
	Rmax&\textless 1s&\multicolumn{3}{c|}{\textbf{\textless 1s}}&34s\\
	\hline
	\multirow{2}{*}{network-prio-2-8-20}&534.44&\textbf{534.44}&\textbf{534.44}&ERROR&TO\\
	&2&\textbf{1}&\textbf{1}&\textbf{1}&\\
	Rmax&1s&\textbf{
	41s}&
	49s&TO&\\
	\hline
	\multirow{2}{*}{stand-tiger-95}&50.38&\textbf{20.41}&\textbf{20.41}&\textbf{20.41}&2.99\\
	&15&4&4&4&\textbf{2}\\
	Rmax&\textless 1s&\multicolumn{3}{c|}{\textbf{\textless 1s}}&364s\\
	\hline
	
\end{longtable}

\begin{table}
	
\centering
	\caption{\textbf{Results on Category B Benchmarks}.
	Comparison to PAYNT on value, size, and time (in that order) on selected benchmarks. The reported time for our approach includes the time of \tool{STORM} for producing the strategy table \textbf{and} the time for learning the FSC.
	Our approach is resulting in similar results as \tool{PAYNT}. Either our value is close to the provided value of \tool{PAYNT} or we outperform it in runtime by at least two orders of magnitude.}
	\label{tab:full-category-B}
\begin{tabular}{|c||c||c|c|c|c|}
	\hline
	&&\multicolumn{3}{c|}{Learning heuristics}&\\
	Model & \textsc{Storm} & base & H$_1$ & H$_2$ & \textsc{Paynt}\\
	\hline\hline

	\multirow{2}{*}{posterior-awareness}&12.0&12.0&12.0&12.0&\textbf{11.99}\\
	&5&\textbf{4}&\textbf{4}&\textbf{4}&\textbf{4}\\
	Rmin&\textless 1s&\multicolumn{3}{c|}{\textbf{\textless 1s}}&\textbf{\textless 1 s}\\
	\hline
	\multirow{2}{*}{problem-paynt}&3.67&\textbf{3.67}&\textbf{3.67}&\textbf{3.67}&\textbf{3.67}\\
	&5&\textbf{3}&\textbf{3}&\textbf{3}&\textbf{3}\\
	Rmin&\textless 1s&\multicolumn{3}{c|}{\textbf{\textless 1s}}&1s\\
	\hline
	\multirow{2}{*}{grid-large-10-5}&40.48&40.48&inf&inf&\textbf{38.72}\\
	&12&8&8&6&\textbf{2}\\
	Rmin&\textless 1s&\multicolumn{3}{c|}{\textbf{\textless 1s}}&10s\\
	\hline

	\multirow{2}{*}{4x5x2-95}&1.29&1.29&1.28&1.26&\textbf{2.02}\\
	&26&18&18&16&\textbf{4}\\
	Rmax&\textless 1s&\multicolumn{3}{c|}{\textbf{\textless 1s}}&2807s\\
	\hline
	\multirow{2}{*}{web-mall}&5.67&5.67&5.67&5.67&\textbf{6.9}\\
	&9&\textbf{3}&\textbf{3}&\textbf{3}&4\\
	Rmax&\textless 1s&\multicolumn{3}{c|}{\textbf{\textless 1s}}&2806s\\
	\hline
	\multirow{2}{*}{query-s3}&415.66&415.66&412.67&345.91&\textbf{486.78}\\
	&124&16&16&8&\textbf{2}\\
	Rmax&\textless 1s&\multicolumn{3}{c|}{\textbf{\textless 1s}}&171s\\
	\hline
	
	\multirow{2}{*}{network}&160.09&160.09&157.92&76.96&\textbf{280.33}\\
	&19&5&6&4&\textbf{3}\\
	Rmax&\textless 1s&\multicolumn{3}{c|}{\textbf{\textless 1s}}&374s\\
	\hline
	
	\multirow{2}{*}{4x3-95}&0.56&0.56&0.76&0.15&\textbf{1.75}\\
	&28&9&9&6&\textbf{3}\\
	Rmax&\textless 1s&\multicolumn{3}{c|}{\textbf{\textless 1s}}&64s\\
	\hline

\end{tabular}

\end{table}

\begin{table}
\centering
	\caption{\textbf{Results on Category C Benchmarks}.
		Comparison to \tool{PAYNT} on value, size and time (in that order). The reported time for our approach includes the time of \tool{STORM} for producing the strategy table \textbf{and} the time for learning the FSC.
		Our approach performs worse than \tool{PAYNT}, but still equally or better than the provided strategy of \tool{STORM}.}
\begin{tabular}{|c||c||c|c|c|c|}
	\hline
	&&\multicolumn{3}{c|}{Learning heuristics}&\\
	Model & \textsc{Storm} & base & H$_1$ & H$_2$ & \textsc{Paynt}\\
	\hline\hline

	\multirow{2}{*}{query-s2}&395.66&395.66&391.9&343.94&\textbf{486.69}\\
	&43&9&9&4&\textbf{2}\\
	Rmax&\textless 1s&\multicolumn{3}{c|}{\textbf{\textless 1s}}&5s\\
	\hline
	
	\multirow{2}{*}{drone-4-1}&0.75&TO&TO&TO&\textbf{0.87}\\
	&3217&&&&\textbf{1}\\
	Pmax&1s&&&&\textbf{2250s}\\
	\hline
	\multirow{2}{*}{lanes-100-combined-new}&19936.52&TO&TO&TO&\textbf{10241.94}\\
	&2705&&&&\textbf{1}\\
	Rmin&70s&&&&\textbf{3s}\\
	\hline

\end{tabular}

\end{table}

\begin{table}
	\caption{Our approach and \tool{PAYNT}, both times out on these benchmarks}
	\centering
	\begin{tabular}{|c||c||c|c|c|c|}
		\hline
		&&\multicolumn{3}{c|}{Learning heuristics}&\\
		Model & \textsc{Storm} & base & H$_1$ & H$_2$ & \textsc{Paynt}\\
		\hline\hline

		\multirow{2}{*}{drone-4-2}&0.86&TO&TO&TO&TO\\
		&2518&\textbf{}&\textbf{}&\textbf{}&\textbf{}\\
		Pmax&\textless 1s&\textbf{}&\textbf{}&\textbf{}&\textbf{}\\
		\hline
		\multirow{2}{*}{drone-8-2}&0.63&TO&TO&TO&TO\\
		&22292&\textbf{}&\textbf{}&\textbf{}&\textbf{}\\
		Pmax&40s&\textbf{}&\textbf{}&\textbf{}&\textbf{}\\
		\hline
		\multirow{2}{*}{network-2-8-20}&6.56&TO&TO&TO&TO\\
		&871&\textbf{}&\textbf{}&\textbf{}&\textbf{}\\
		Rmin&\textless 1s&\textbf{}&\textbf{}&\textbf{}&\textbf{}\\
		\hline
		\multirow{2}{*}{network-3-8-20}&11.91&TO&TO&TO&TO\\
		&1849&\textbf{}&\textbf{}&\textbf{}&\textbf{}\\
		Rmin&4s&\textbf{}&\textbf{}&\textbf{}&\textbf{}\\
		\hline
	\end{tabular}
	
\end{table}

\end{small}

\begin{table}[t]
	\caption{Data for models where \textsc{PAYNT} times out.  `-' denotes an error while evaluating the FSC. Columns contain: (1) the computed property value and time taken, and
		(2) the size of input Markov chain or resulting FSC, resp.}
	\makebox[\textwidth][c]{
		\resizebox{1.1\textwidth}{!}{
			\begin{tabular}{|c||c|c||c|c|c|c|c|c| |c||c|c||c|c|c|c|c|c|}
	\hline
	&\multicolumn{2}{c||}{}&\multicolumn{6}{c||}{Learning heuristics}&&\multicolumn{2}{c||}{ }&\multicolumn{6}{c|}{Learning heuristics}\\
	Model & \multicolumn{2}{c||}{\textsc{Storm}} & \multicolumn{2}{c|}{ base} &  \multicolumn{2}{c|}{H$_1$} &  \multicolumn{2}{c||}{H$_2$}&
	Model & \multicolumn{2}{c||}{\textsc{Storm}} & \multicolumn{2}{c|}{base} & \multicolumn{2}{c|}{H$_1$} &  \multicolumn{2}{c|}{H$_2$} \\
	\hline\hline
	{grid-large-20-5}&80.51&$<$1s&80.51&$<$1s&80.51&$<$1s&-&$<$1s&
	{refuel-20}&0.14&73s&0.14&75s&0.23&74s&0.23&74s\\
	Rmin&\multicolumn{2}{c||}{2}&\multicolumn{2}{c|}{1}&\multicolumn{2}{c|}{1}&\multicolumn{2}{c||}{1}&
	Pmax&\multicolumn{2}{c||}{46}&\multicolumn{2}{c|}{4}&\multicolumn{2}{c|}{4}&\multicolumn{2}{c|}{3}\\
	\hline
	{grid-large-30-5}&120.52&$<$1s&120.52&$<$1s&120.52&$<$1s&-&$<$1s&
	{rocks-12}&38.0&1s&38.0&13s&38.0&10s&inf&10s\\
	Rmin&\multicolumn{2}{c||}{2}&\multicolumn{2}{c|}{1}&\multicolumn{2}{c|}{1}&\multicolumn{2}{c||}{1}&
	Rmin&\multicolumn{2}{c||}{9}&\multicolumn{2}{c|}{2}&\multicolumn{2}{c|}{2}&\multicolumn{2}{c|}{1}\\
	\hline
	{hallway2}&0.05&9s&0.05&15s&0.16&15s&0.06&13s&
	{rocks-16}&44.0&2s&44.0&34s&44.0&38s&inf&36s\\
	Rmax&\multicolumn{2}{c||}{4517}&\multicolumn{2}{c|}{16}&\multicolumn{2}{c|}{17}&\multicolumn{2}{c||}{12}&
	Rmin&\multicolumn{2}{c||}{9}&\multicolumn{2}{c|}{2}&\multicolumn{2}{c|}{2}&\multicolumn{2}{c|}{1}\\
	\hline
	network-prio.2.8.20&534.44&1s&534.44&41s&534.44&49s&-&48s&\multicolumn{8}{c}{}\\
	Rmax&\multicolumn{2}{c||}{2}&\multicolumn{2}{c|}{1}&\multicolumn{2}{c|}{1}&\multicolumn{2}{c||}{1}&\multicolumn{9}{c}{}\\
	\cline{1-9}
\end{tabular}

}}
	
	\label{tab:good-examples}
\end{table}

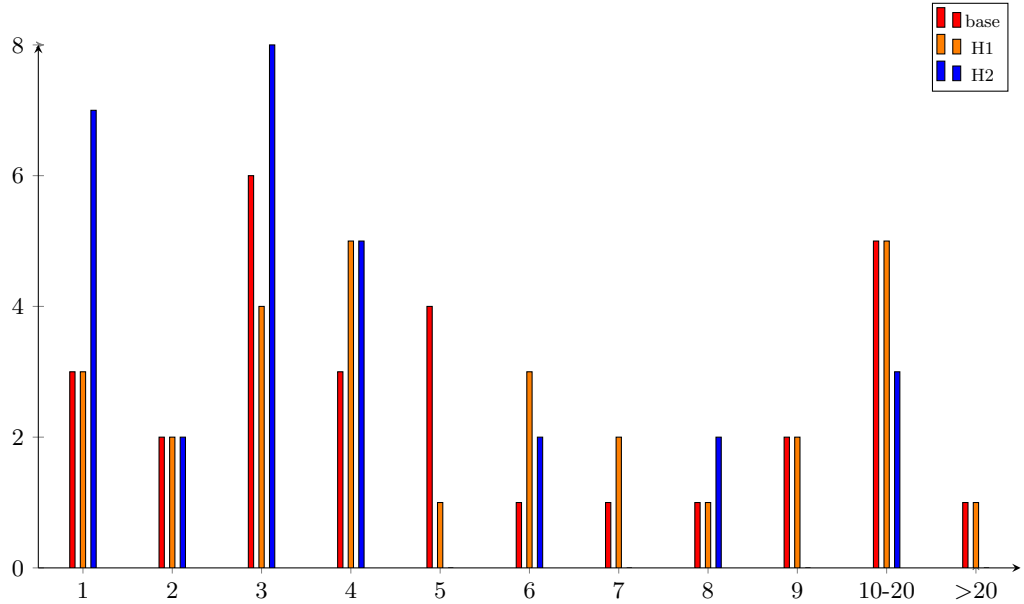
\begin{figure}[t]

	\centering	
	\begin{tikzpicture}[scale=1]
			\begin{axis}[
				ybar,
				width=1.2\textwidth,
				height=0.7\textwidth,
				bar width=2pt,
				xtick=data,
				axis x line=bottom,
				axis y line=left,
				legend style={nodes={scale=0.75, transform shape},inner sep=1.5pt, xshift=1mm, yshift=7mm},
				xmin =0.5,
				xmax =11.5,
				xticklabels={1,2,3,4,5,6,7,8,9,10-20,$>$20}]
				\pgfplotstableread{data/sizes-hist.csv}\loadedtable;
				\addplot[ybar,fill=color1] table[x=size,y=num-cutoff] {\loadedtable};
				\addplot[ybar,fill=color2] table[x=size,y=num-distribution] {\loadedtable};
				\addplot[ybar,fill=color3] table[x=size,y=num-minimize] {\loadedtable};
				\legend{base,H1,H2};
				\end{axis}
		\end{tikzpicture}
	\caption{Historgram of the size of our learned FSC.}
	\label{fig:fsc-sizes}

\end{figure}

\end{document}